\documentclass[11pt]{article}

\usepackage{microtype}
\usepackage{graphicx}
\usepackage{booktabs}
\usepackage{amsmath}
\usepackage{amsfonts}

\usepackage{amssymb}
\usepackage{bm}
\usepackage{enumitem}
\usepackage{xspace}
\usepackage{makecell}
\usepackage{multirow}
\usepackage{wrapfig}
\usepackage{algorithm}
\usepackage{algpseudocode}
\usepackage{ziplab-tech-report}
\usepackage{hyperref}
\usepackage[capitalize]{cleveref}

\newcommand{\reporttitle}{VolSplat: Rethinking Feed-Forward 3D Gaussian Splatting with Voxel-Aligned Prediction}
\newcommand{\reportshorttitle}{VolSplat}
\newcommand{\reportsup}[1]{\textsuperscript{\textit{\ensuremath{#1}}}}
\newcommand{\reportauthors}{%
Weijie Wang\reportsup{1,2,*} \hspace{0.7em}
Yeqing Chen\reportsup{1,*} \hspace{0.7em}
Zeyu Zhang\reportsup{2} \hspace{0.7em}
Hengyu Liu\reportsup{2,3}\\[-0.1em]
Haoxiao Wang\reportsup{1} \hspace{0.7em}
Zhiyuan Feng\reportsup{4} \hspace{0.7em}
Wenkang Qin\reportsup{2} \hspace{0.7em}
Feng Chen\reportsup{5}\\[-0.1em]
Jia-Wang Bian\reportsup{6} \hspace{0.7em}
Zheng Zhu\reportsup{2,\ddagger} \hspace{0.7em}
Donny Y. Chen\reportsup{7} \hspace{0.7em}
Bohan Zhuang\reportsup{1,\ddagger}}
\newcommand{\reportaffiliations}{%
\reportsup{1} Zhejiang University \quad
\reportsup{2} GigaAI \quad
\reportsup{3} The Chinese University of Hong Kong \quad
\reportsup{4} Tsinghua University\\
\reportsup{5} Adelaide University \quad
\reportsup{6} Nanyang Technological University \quad
\reportsup{7} Monash University}
\newcommand{\reportdate}{September 23, 2025}
\newcommand{\reportconference}{The 19th European Conference on Computer Vision (ECCV 2026)}

\newcommand{\reportproject}{\href{https://lhmd.top/volsplat}{lhmd.top/volsplat}}
\newcommand{\reportemail}{\href{mailto:zhengzhu@ieee.org}{zhengzhu@ieee.org} and \href{mailto:bohan.zhuang@zju.edu.cn}{bohan.zhuang@zju.edu.cn}}

\newcommand{\reportkeywords}{3D Gaussians, Feed-Forward Reconstruction, View Synthesis}
\newcommand{\reportfootnote}{\reportsup{*} Equal contribution. \reportsup{\ddagger} Corresponding authors.}
\newcommand{\reportpdfauthors}{Weijie Wang, Yeqing Chen, Zeyu Zhang, Hengyu Liu, Haoxiao Wang, Zhiyuan Feng, Wenkang Qin, Feng Chen, Jia-Wang Bian, Zheng Zhu, Donny Y. Chen, Bohan Zhuang}

\techreportlabel{ZIP Lab Technical Report}
\techreportshorttitle{\reportshorttitle}
\techreportdate{\reportdate}
\techreportconference{\reportconference}
\techreportcode{}
\techreportproject{\reportproject}
\techreportemail{\reportemail}
\techreportorcid{}
\techreportkeywords{\reportkeywords}
\techreportfootnote{\reportfootnote}
\techreportsetlogoheight{0.82cm}
\techreportlogos{%
  \techreportlogo{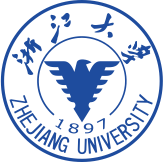}
  \techreportlogo{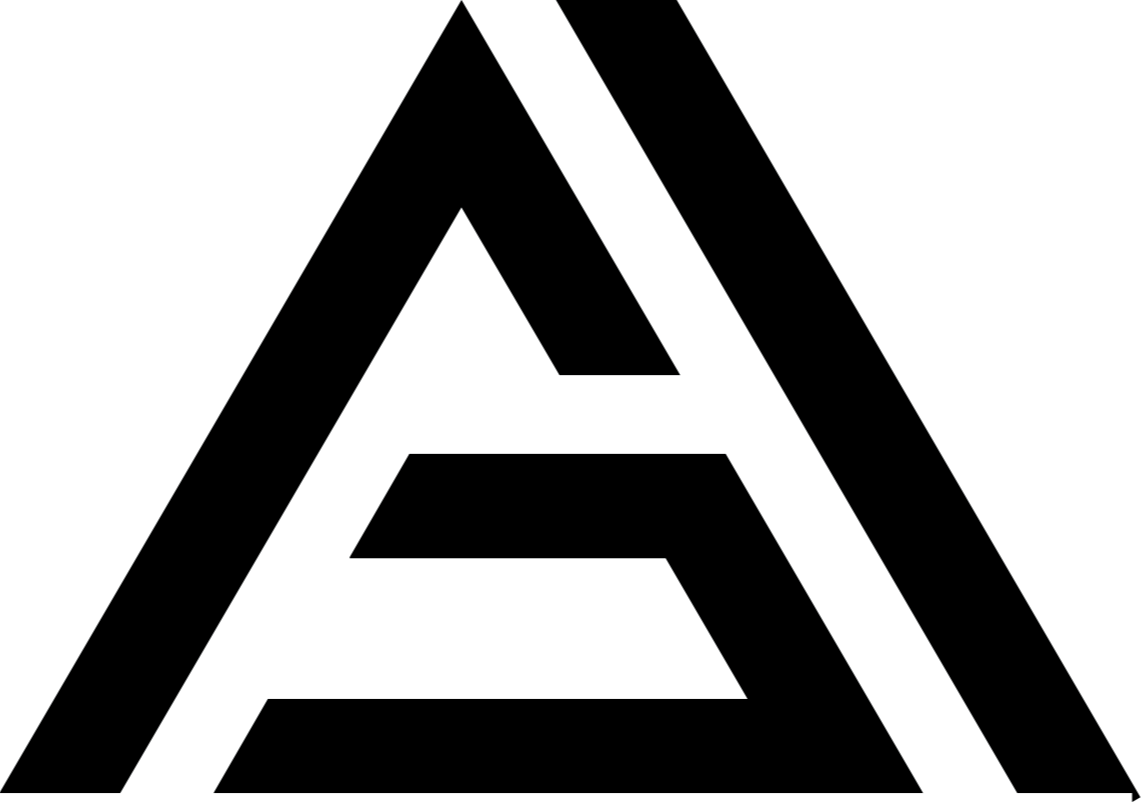}
  \techreportlogo{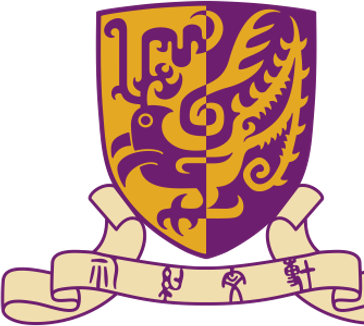}
  \techreportlogo{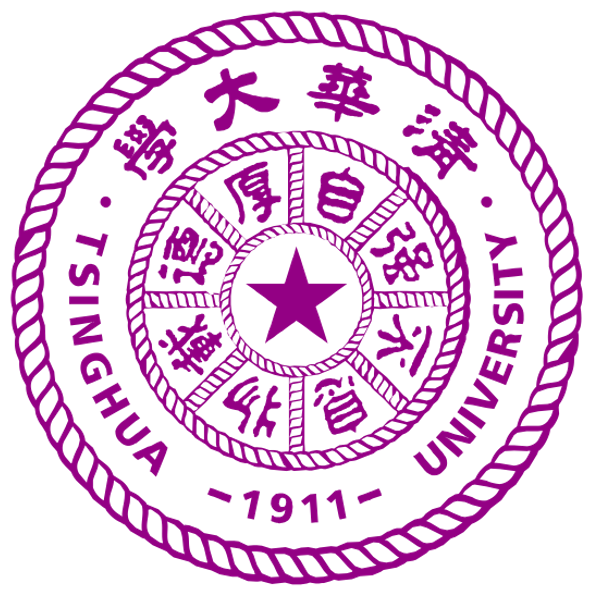}
  \techreportlogo{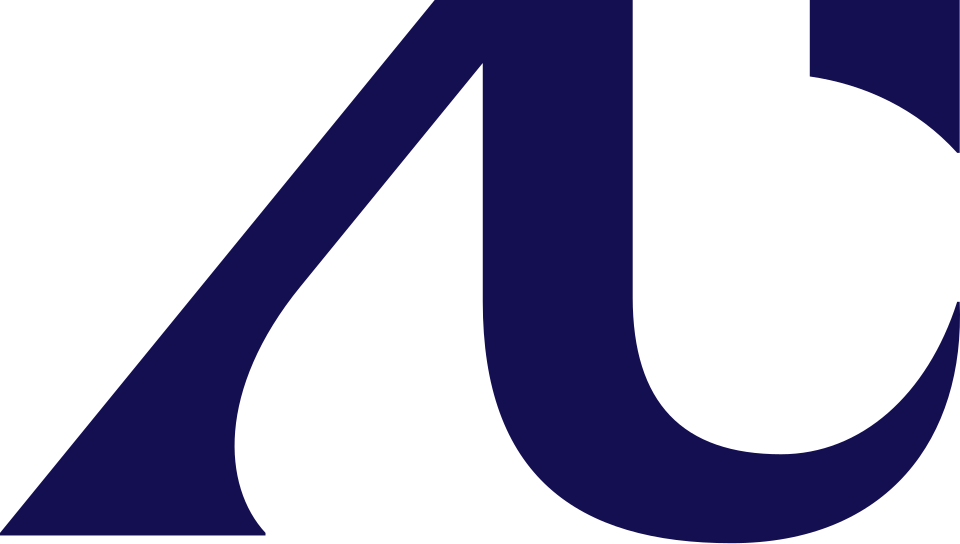}
  \techreportlogo{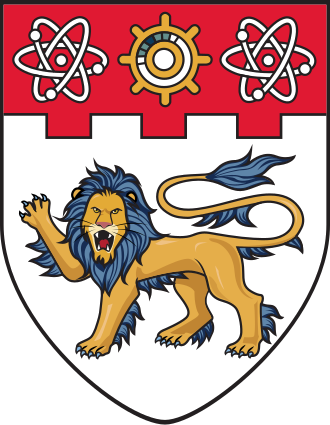}
  \techreportlogo{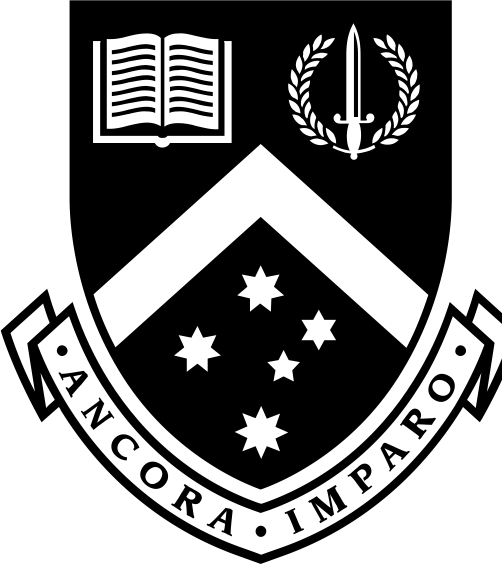}
}

\hypersetup{
  colorlinks=true,
  linkcolor=ziplabblue,
  citecolor=ziplabblue,
  urlcolor=ziplabblue,
  pdftitle={\reporttitle},
  pdfauthor={\reportpdfauthors},
  pdfsubject={arXiv preprint},
  pdfkeywords={\reportkeywords}
}

\newcommand{\method}{VolSplat\xspace}
\newcommand{\boldstart}[1]{\vspace{0.1in}\noindent\textbf{#1}}
\newcommand\rurl[1]{\href{https://#1}{\nolinkurl{#1}}}
\crefname{section}{Sec.}{Secs.}
\crefname{figure}{Fig.}{Figs.}
\crefname{table}{Tab.}{Tabs.}
\crefname{appendix}{Appendix}{Appendices}

\begin{document}

\ziplabmaketitle{\reporttitle}{\reportauthors}{\reportaffiliations}{%
  \begin{center}
  \begin{minipage}{\textwidth}
    \centering
    \vspace{0.08cm}
    \includegraphics[width=\textwidth]{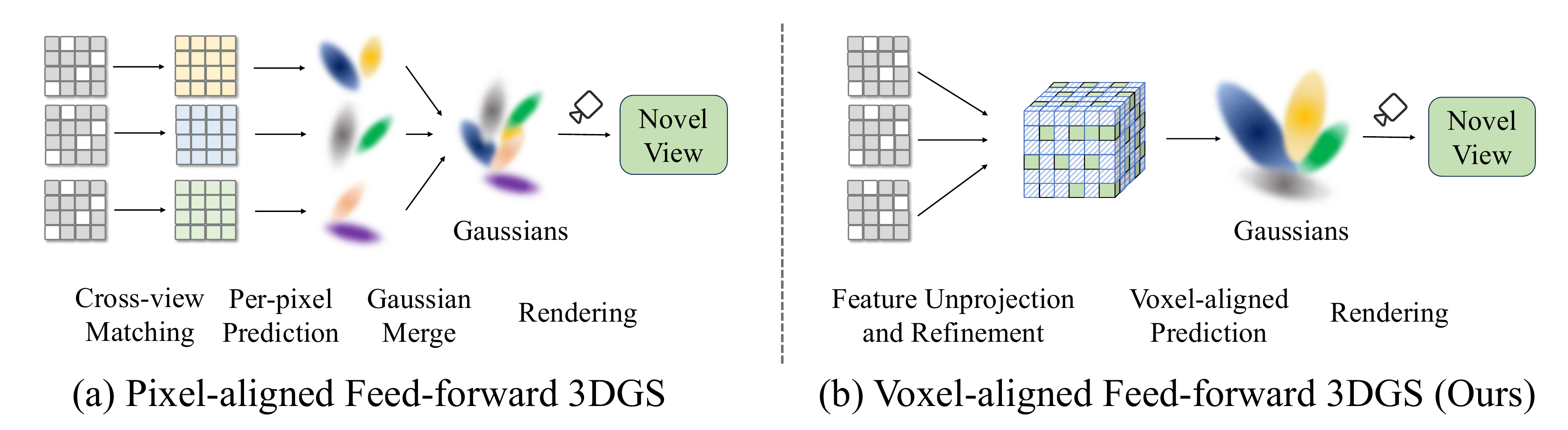}
    \vspace{-0.85em}
    {\captionsetup{font=scriptsize,aboveskip=2pt,belowskip=0pt}
    \captionof{figure}{\textbf{Comparison between the pixel-aligned feed-forward method and our approach.}
    Pixel-aligned feed-forward 3DGS methods suffer from two primary limitations: 1) 2D feature matching struggles to effectively resolve the multi-view alignment problem, and 2) the Gaussian density is constrained and cannot be adaptively controlled according to scene complexity. We propose \method, a framework that directly regresses Gaussians from 3D features based on a voxel-aligned prediction strategy.}
    \label{fig:teasor}}
  \end{minipage}
  \end{center}
}
\ziplabendtitle{%
Feed-forward 3D Gaussian Splatting (3DGS) has emerged as a highly effective solution for novel view synthesis. Existing methods predominantly rely on a \emph{pixel-aligned} Gaussian prediction paradigm, where each 2D pixel is mapped to a 3D Gaussian. We rethink this widely adopted formulation and identify several inherent limitations: it renders the reconstructed 3D models heavily dependent on the number of input views, leads to view-biased density distributions, and introduces alignment errors, particularly when source views contain occlusions or low texture. To address these challenges, we introduce \method, a new multi-view feed-forward paradigm that replaces pixel alignment with voxel-aligned Gaussians. By directly predicting Gaussians from a predicted 3D voxel grid, it overcomes pixel alignment's reliance on error-prone 2D feature matching, ensuring robust multi-view consistency. Furthermore, it enables adaptive control over density based on 3D scene complexity, yielding more faithful Gaussians, improved geometric consistency, and enhanced novel-view rendering quality. Experiments on widely used benchmarks demonstrate that \method achieves state-of-the-art performance, while producing more plausible and view-consistent results.}

\clearpage
\section{Introduction}

3D reconstruction is a cornerstone of modern robotics, empowering autonomous systems with the critical ability to perceive, map, and comprehend their physical environment~\cite{wang2025transdiff}, which is fundamental for advanced navigation, object manipulation, and world models.
Traditional optimization based approaches, including Neural Radiance Fields (NeRF)~\cite{mildenhall2021nerf} and 3D Gaussian Splatting (3DGS)~\cite{kerbl20233d}, obtain high fidelity results by iteratively enforcing photometric or geometric consistency. These methods achieve excellent accuracy but are computationally intensive and slow to run at inference time. By contrast, feed-forward approaches~\cite{yu2021pixelnerf, wang2021ibrnet, chen2021mvsnerf, charatan2024pixelsplat, zhang2024gs, chen2024mvsplat, xu2025depthsplat, wang2025zpressor, wang2024freesplat, chen2024mvsplat360, kang2025ilrm} trade per instance optimization for fast learned inference. A single forward pass predicts scene geometry or a 3D representation directly from input images. This speed and simplicity make feed-forward systems attractive for real time applications, large scale datasets, and downstream tasks that require many reconstructions.

Prior feed-forward 3DGS methods~\cite{charatan2024pixelsplat, chen2024mvsplat, xu2025depthsplat, wang2024freesplat, kang2025ilrm, wang2025drivegen3d, wang2026feed} commonly rely on pixel alignment as their fundamental mechanism for associating image features with pixel aligned Gaussians. In this design, per-pixel features from precomputed image feature maps are unprojected to define the corresponding Gaussians. The prevailing consensus has been to perform fusion directly within the 2D feature representation. However, pixel aligned designs inherit two primary limitations. 
1) Sampling at discrete pixel locations is sensitive to camera calibration and discretization error, produces inconsistent sampling patterns across views.
2) The rigid pixel-to-Gaussian association enforces a uniform density distribution that ignores scene complexity, leading to redundant primitives in simple regions while failing to capture fine-grained 3D structures.

In this work, we shift the alignment paradigm from pixels to voxels, as illustrated in \cref{fig:teasor}. Instead of sampling features at projected pixel coordinates, we align and aggregate image features directly into a 3D voxel grid. Multi-view image features are aggregated directly into this 3D voxel space, effectively decoupling feature fusion from the camera view frustums. Within this unified voxel space, we employ a 3D U-Net~\cite{cciccek20163d} to reason about scene geometry and appearance volumetrically. Finally, rather than predicting Gaussians per pixel, we predict primitives directly from the refined voxel features, allowing the distribution of 3D Gaussians to be determined by the volumetric structure itself.

There are practical and conceptual advantages to voxel alignment. Specifically, volumetric aggregation reduces floaters and view dependent inconsistency because information from multiple views is fused into a shared 3D container before Gaussian prediction. Simultaneously, operating in a 3D grid enables the use of well studied 3D decoder and regularization strategies, which naturally encode locality and geometrical context. Instead of the integration of auxiliary 3D signals such as depth maps~\cite{xu2025depthsplat} and point clouds~\cite{shi2025pmloss}, our approach naturally resolves spatial ambiguities within the unified voxel space, thereby eliminating the need for such ad hoc priors or auxiliary supervision signals. Furthermore, voxel representations are amenable to modern acceleration strategies such as sparse data structures, making the approach practical at the resolutions required for high quality reconstruction.

\begin{figure}[t]
    \centering
    \includegraphics[width=\textwidth]{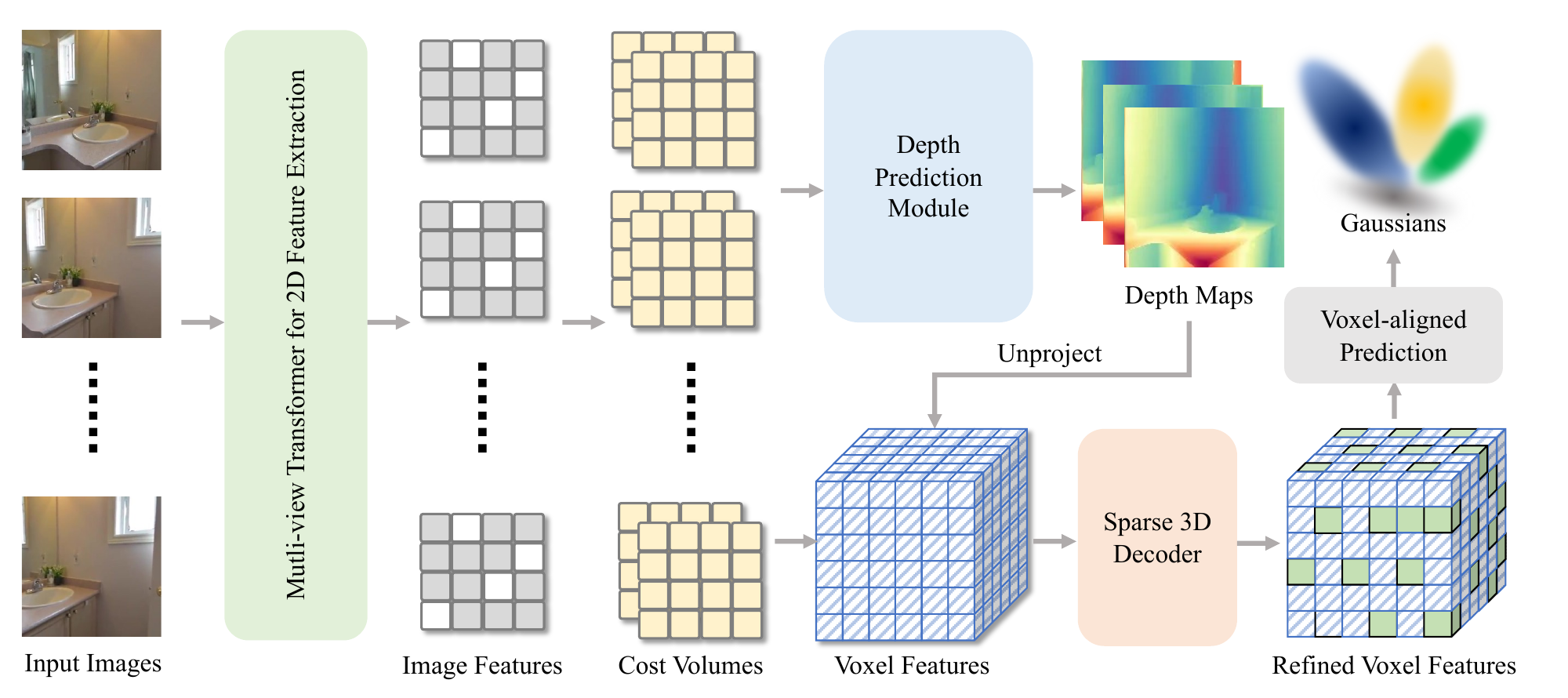}
    \caption{\textbf{Overview of VolSplat.} Given multi-view images as input, we first extract 2D features for each image using a Transformer-based network and construct per-view cost volumes with plane sweeping. Depth Prediction Module then estimates a depth map for each view, which is used to unproject the 2D features into 3D space to form a voxel feature grid. Subsequently, we employ a sparse 3D decoder (details in Sec.~\ref{sec:refinement}) to refine these features in 3D space and predict the parameters of a 3D Gaussian for each occupied voxel. Finally, novel views are rendered from the predicted 3D Gaussians.}
    \label{fig:pipeline}
    \vspace{-0.3cm}
\end{figure}

In this paper we present a feed-forward three dimensional reconstruction framework built around voxel alignment. As shown in \cref{fig:pipeline}, we first construct 3D feature grids using the extracted 2D image features, then refine the 3D features and use them to predict voxel-aligned Gaussians. We analyze the alignment errors that arise in pixel aligned pipelines and show how voxel alignment reduces these errors both conceptually and empirically. Through systematic experiments on synthetic and real world benchmarks, we demonstrate that voxel aligned feed-forward models achieve more accurate and robust reconstructions than comparable pixel aligned baselines on large-scale benchmarks such as RealEstate10K~\cite{zhou2018stereo}, ScanNet~\cite{dai2017scannet} and ACID~\cite{liu2021infinite}. Our contributions are as follows:

\begin{itemize}[itemsep=2pt]
    \item We introduce voxel alignment as a principled alternative to pixel alignment for feed-forward 3DGS and present a practical end-to-end framework.
    \item We provide an analysis of alignment induced errors in pixel aligned systems and show how volumetric aggregation mitigates these failure modes.
    \item Experimental results %
    demonstrate that
    VolSplat achieves state-of-the-art (SOTA) performance on several large-scale benchmarks.
\end{itemize}

\section{Related Work}

\boldstart{Novel view synthesis.}
Traditional approaches to Novel View Synthesis (NVS) primarily rely on geometry-based rendering methods that reconstruct explicit 3D scene geometry from images~\cite{debevec2023modeling}, image-based rendering techniques that interpolate between captured views without full 3D reconstruction~\cite{ji2017deep}, and light field rendering that samples and reprojects densely captured rays in space~\cite{levoy2023light}. These methods required either accurate geometric proxies, densely sampled viewpoints, or both to produce convincing visual results, limiting their applicability in real-world scenarios.
The emergence of NeRF~\cite{mildenhall2021nerf} marked a paradigm shift, significantly improving both rendering quality and robustness over prior methods, which learns a continuous, implicit scene representation by utilizing a MLP to map position and viewing direction to a corresponding color and volume density.
While NeRF-based methods~\cite{barron2021mip, barron2023zip} require a long training time due to the per-ray rendering.
3DGS~\cite{kerbl20233d} and its variants~\cite{fan2024lightgaussian,zhu2024fsgs,liu2025flexgs} have been introduced to represent the 3D scene using a set of anisotropic 3D Gaussians.

\boldstart{3D voxelization.}
Voxelization, which discretizes 3D space into regular voxel grids, has been a foundational representation in 3D reconstruction and modeling~\cite{meagher1982geometric}. Prior methods used dense grids for their simplicity, but suffered from high memory costs and poor scalability~\cite{kaufman2005overview}. To address this, sparse structures like octrees were introduced for more efficient storage and computation~\cite{koneputugodage2023octree}.
In modern applications, voxels are widely used as input to 3D Convolutional Neural Network (CNN) for tasks such as object detection~\cite{zhou2018voxelnet} and semantic segmentation~\cite{riegler2017octnet,wang2026diffusion}. More recently, voxels are often used as sparse scaffolding rather than as the final representation, supporting more advanced rendering techniques. Representative methods include Plenoxels~\cite{fridovich2022plenoxels} and K-Planes~\cite{fridovich2023k}, which optimize voxel-based radiance fields for fast, high-quality rendering, as well as structured strategies such as Scaffold-GS~\cite{lu2024scaffold} and Octree-GS~\cite{ren2024octree}, which leverage voxel grids to organize and accelerate 3DGS.

\boldstart{Feed-forward 3D Gaussian Splatting.}
Recent developments in feed-forward 3DGS~\cite{charatan2024pixelsplat, chen2024mvsplat, xu2025depthsplat, wang2024freesplat, zhang2024gaussian, jiang2025anysplat,ni2025wonderturbo,huang2025spfsplat,wang2026trisplat} offer a compelling alternative that directly predicts 3D Gaussians from input images in a single forward pass:
pixelSplat~\cite{charatan2024pixelsplat} proposes a two-view feed-forward pipeline that combines epipolar transformers and depth prediction to generate Gaussians. MVSplat~\cite{chen2024mvsplat} introduces a cost-volume-based fusion strategy to enhance multi-view consistency. DepthSplat~\cite{xu2025depthsplat} leverages monocular depth features to improve fine 3D structure reconstruction from sparse views. 
Follow-up work extends feed-forward 3DGS to more complex scenarios, including pose-free inputs~\cite{ye2024no, kang2025selfsplat}, online stream inputs~\cite{wang2024freesplat}, and more dense inputs~\cite{wang2025zpressor}, which can provide priors for the world models~\cite{zhang2026panflow,wang2026world,wang2026latent}. 
While these works adopt a pixel-aligned strategy to predict Gaussian primitives, the pixel-wise formulation struggles to handle multiple input views due to redundancy and inconsistency across pixels. 
Existing methods attempt to improve the per-pixel strategy by pruning the number of Gaussians~\cite{zhang2024gaussian}, token merging~\cite{ziwen2024long,wang2025zpressor} and voxel-based fusion~\cite{jiang2025anysplat,wang2025learning,liu2025worldmirror}.
However, these approaches do not fundamentally address the limitations inherent in per-pixel processing. 
EVolSplat~\cite{miao2025evolsplat} has explored voxel features in autonomous driving scenarios, but it has not been generalized to general scenarios and requires explicit 3D point clouds as intermediate representations.
In contrast, our method introduces a voxel-aligned method, which eliminate the need for per-query 2D prediction patterns. This alignment enables more stable multi-view fusion, cleaner occlusion handling, and more coherent joint inference of geometry and appearance.

\section{Method}

\subsection{Preliminary and Observation}
Feed-forward 3D reconstruction aims to learn a mapping from $N$ input images $\mathcal{I}=\{{\mathbf I}_{i}\}_{i=1}^N$ where ${\mathbf I}_i \in \mathbb{R}^{H \times W \times 3}$ and their corresponding camera poses $\mathcal{P}=\{{\mathbf P}_{i}\}_{i=1}^N$, to a 3D scene representation. In the context of pixel-aligned 3DGS, features are extracted from images and refined by cross-view interaction:
\begin{align}
    \label{eq:feature_extraction}
    \mathcal{F}=\{\mathbf{F}_i\}_{i=1}^N=h(\Phi_\mathrm{image}(\mathcal{I}, \mathcal{P})), \quad \mathbf{F}_i
    \in\mathbb{R}^{\frac{H}{p}\times \frac{W}{p}\times C}
\end{align}
where $\Phi_\mathrm{image}$ is a pretrained image encoder. The function $h$ is responsible for processing these features from different viewpoints, with its core purpose being to perform cross-view feature matching and fusion. For pixel-aligned Gaussian prediction, the features must be upsampled to the same resolution as the input image:
\begin{align}
    \mathcal{F}_\mathrm{full} = U(\mathcal{F}) \in \mathbb{R}^{N \times H \times W \times C},
\end{align}
where $\mathcal{F}_\mathrm{full}$ denotes the full-resolution feature maps, $U$ is a feature upsampler such as CNN-based network (in MVSplat~\cite{chen2024mvsplat}) and deconvolution-based network~\cite{odena2016deconvolution} (in DepthSplat~\cite{xu2025depthsplat}). Per-pixel Gaussian predictions are then performed using the upsampled features:
\begin{align}
    \mathcal{G} = \left\{(\mathbf{\mu}_{i}, \mathbf{\Sigma}_{i}, \mathbf{\alpha}_{i}, \mathbf{c}_{i}\right)\}_{i=1}^{H \times W \times N} = \Psi_\mathrm{pred}(\mathcal{F}_\mathrm{full}, \mathcal{P}),
\end{align}
where the position of the Gaussians are determined by the predicted depth and pixel location.

While straightforward, this pixel-aligned formulation introduces two critical limitations. 
First, the geometric accuracy of the reconstruction is critically dependent on the quality of the predicted depth map. After depth unprojecting features into 3D space, the lack of interaction with neighboring points within the 3D space significantly contributes to the generation of floaters. 
Second, the structure of the 3D representation is rigidly tied to the 2D image grid. The total number of Gaussians is fixed at $|S|=H\times W\times N$, which is often suboptimal and cause an over-densification of Gaussians on simple, texture-less surfaces and an insufficient number for representing complex geometry not captured at the pixel level. These observations reveal a fundamental bottleneck and motivate our proposed voxel-aligned framework, designed to decouple the 3D representation from the 2D pixel grid.

\subsection{3D Feature Construction}
\noindent\textbf{Feature extraction and matching.}
For $N$ input images, we first apply a weight-sharing ResNet~\cite{he2016deep} backbone to each RGB image to obtain $p\times$ downsampled feature maps. These features are then refined with cross-view attention that exchanges information with the two nearest neighboring views. For efficiency, this cross-attention is implemented with the local window attention~\cite{liu2021swin}. After this stage we obtain cross-view-aware Transformer features $\left\{ {F}_i \right\}_{i=1}^N ({F}^i \in \mathbb{R}^{\frac{H}{p} \times \frac{W}{p} \times C})$ , where $C$ denotes the feature dimension. 

Next, we build per-view cost volumes $\left\{ C_i \right\}_{i=1}^N$ using a plane-sweep strategy~\cite{xu2023unifying}. For each view $i$, we sample $D$ candidate depths $\left\{ d_m \right\}_{m=1}^D$, warp the feature from neighboring views to the reference view at each hypothesized depth, and compute pairwise feature similarities~\cite{chen2024mvsplat}.These similarities are aggregated by dot-product matching and stacked along the depth axis to form $\left\{ {C}_i \right\}_{i=1}^N$, where $C_i \in \mathbb{R}^{\frac{H}{p} \times \frac{W}{p} \times D}$.

To produce robust, multi-view consistent depth estimates, a depth module fuses the monocular features $\left\{ {F}^i_{mono} \right\}_{i=1}^N ({F}^i_{mono} \in \mathbb{R}^{\frac{H}{p} \times \frac{W}{p} \times C})$ with the cost volume ${C}^i$ and regresses a dense per-pixel depth map $D_i \in \mathbb{R}^{H \times W }$, which serves as a geometric prior for lifting image features into 3D space. These per-view features ${F}^i$ and depths $D_i$ are used in the next stage to construct 3D point clouds and voxel-based features for volumetric reasoning.

\noindent\textbf{Lifting to 3D feature.}
Given the predicted depth maps  $D_i$  and camera parameters, we conveniently aggregate different depth map views by transforming the point clouds into a global coordinate system. First each pixel $(u, v)$ in image space is unprojected to a 3D point in the camera coordinate frame using the camera intrinsics. Then the 3D point is transformed into the world coordinate system via the corresponding extrinsic parameters, including the rotation matrix $R_i$ and translation $T_i$ vector. 
\begin{equation}
    P_{\mathrm{world}}
    = R_i\,P_{\mathrm{cam}} + T_i 
    = R_i\!\left(D_i(u,v)\,K^{-1}\begin{bmatrix}u\\ v\\ 1\end{bmatrix}\right) + T_i.
    \label{eq:unproject}
\end{equation}
By repeating this process across all views, we obtain a dense $|S|=H \times W \times N$ point cloud in world space, where each 3D point is associated with its corresponding image feature.

\begin{figure}[t]
\centering
\includegraphics[width=\linewidth]{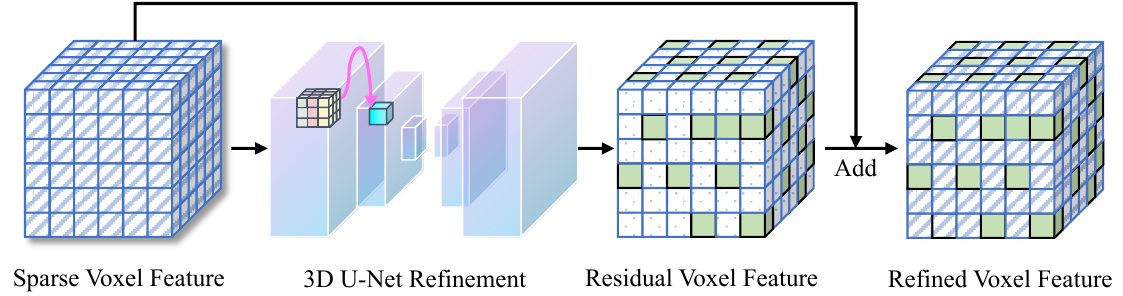}
\vspace{-0.5cm}
\caption{\textbf{Architecture of Sparse 3D decoder.}  Sparse 3D features are fed into a 3D U-Net for processing, which predicts residual features for each voxel. These residual features are then added to the original 3D voxel features to obtain the refined features.}
\label{fig:refinement}
\vspace{-0.5cm}
\end{figure}
To convert the unstructured dense point cloud $P$ into a structured volumetric representation, we voxelize the points~\cite{wang2024embodiedscan}. For each 3D point $p=(x_p,y_p,z_p)$ we compute integer voxel index $(i,j,k)$ by dividing by the voxel size $v_s$ and rounding. 
\begin{equation}
    i = \operatorname{rnd}\left( \frac{x_p}{v_s} \right),\ j = \operatorname{rnd}\left( \frac{y_p}{v_s} \right),\ k = \operatorname{rnd}\left( \frac{z_p}{v_s} \right),
\end{equation}
where $\operatorname{rnd}(\cdot)$ denotes rounding to the nearest integer.

Let $S_{i,j,k}$ be the set of all points falling into voxel $(i,j,k)$ and $f_{p}$be the image feature corresponding to each point $p \in S_{i,j,k}$ The features within this voxel are aggregated via average pooling along the channel dimension, resulting in the voxel feature $V_{i,j,k}$:
\begin{align}
    V_{i,j,k} &= \frac{1}{\lvert S_{i,j,k}\rvert}\sum_{p\in S_{i,j,k}} f_p.
    \label{eq:voxel_feature}
\end{align}

\subsection{Feature Refinement and 3D Gaussians Prediction}
\noindent\textbf{Feature refinement.}
\label{sec:refinement}
To improve the spatial consistency and structural fidelity of the voxel representation, we apply an explicit voxel feature refinement stage as shown in ~\cref{fig:refinement}. Given an input voxel grid $V$ (with per-voxel feature vectors), a sparse convolutional 3D U-Net~\cite{cciccek20163d} $\mathcal{R}$ predicts a residual voxel field $R$:
    \begin{align}
    R &= \mathcal{R}(V), \quad 
    {R}_{i}\in\mathbb{R}^{\mathcal{V}\times C},
    \label{eq:refine}
\end{align}
where $\mathcal{V}$ denotes the set of occupied voxels
and the refined voxel features are obtained by a residual update:
\begin{align}
    V' &= V + R, \quad 
    {V'}_{i}\in\mathbb{R}^{\mathcal{V}\times C},
    \label{eq:voxel_update}
\end{align}
The refinement network is implemented with hierarchical sparse 3D convolutional blocks, symmetric encoder-decoder stages, and upsampling layers connected by skip connections. This architecture enables multi-scale fusion of local and global geometric context while keeping computation efficient through sparsity. The residual formulation encourages the network to learn correction terms (fine geometric detail and consistency cues) rather than relearning the entire feature content, which empirically stabilizes training and preserves the coarse voxel information supplied by the lifting stage.

\noindent\textbf{3D Gaussians prediction.}
The output of our network for each voxel $v$ is a set of learnable Gaussian parameters $\left\{ [\bar{\mu}_j, \bar{\alpha}_j, {\Sigma}_j, c_j \right ]\in \mathbb{R}^{38}\} $. These include the offset of the Gaussian center $\bar{\mu}_j$, opacity $\bar{\alpha}_j$, covariance ${\Sigma}_j$, and spherical harmonic color representation $c_j$. To obtain the final rendering parameters, we apply the following transformations:
\begin{align}
    \mu_j &= r\cdot(\sigma(\bar{\mu}_j)-0.5) + \mathrm{Center}_j \nonumber,\\
    \alpha_j &= \sigma(\bar{\alpha}_j),
    \label{eq:gaussian_params}
\end{align}
where $\mu_j$ is the predicted 3D Gaussian center, and $\mathrm{Center}_j$ is the centroid of voxel $v$. We utilize the sigmoid activation $\sigma(\cdot)$ to restrict the learnable offset within a localized neighborhood. Specifically, the $-0.5$ shift facilitates symmetrical, bi-directional movement from the voxel center, while $r$ (set to $3 \times$ voxel size) acts as a scaling factor to control the effective spatial extent of these refinements.

\subsection{Training Objectives}
\label{sec:training}
Our network predicts a collection of 3D Gaussians 
$\{(\mu_v, \alpha_v, \Sigma_v, c_v)\}_{v \in \mathcal{V}}$. These per-voxel Gaussians are subsequently used to synthesize images at novel camera poses. To ensure a fair comparison and maintain benchmarking consistency with SOTA feed-forward methods, we follow the training protocol established by DepthSplat~\cite{xu2025depthsplat}. Specifically, the network is trained end-to-end using ground-truth RGB images as supervision. For a forward pass that renders $M$ novel views, we optimize a combined photometric and perceptual loss:
\begin{equation}
    \mathcal{L} = \sum_{m=1}^{M} \left( \mathcal{L}_{\mathrm{MSE}}(I_{\mathrm{render}}^{(m)}, I_{\mathrm{gt}}^{(m)}) + \lambda \mathcal{L}_{\mathrm{LPIPS}}(I_{\mathrm{render}}^{(m)}, I_{\mathrm{gt}}^{(m)}) \right),
    \label{eq:gs_loss}
\end{equation}
where $M$ is the number of novel views rendered in a single pass. Following DepthSplat~\cite{xu2025depthsplat}, the weight $\lambda$ for the perceptual loss $\mathcal{L}_{\mathrm{LPIPS}}$ is set to 0.05.

\section{Experiments}

\subsection{Experimental Setup}
\noindent\textbf{Datasets.}
We train our method using two expansive datasets, RealEstate10K~\cite{zhou2018stereo} and ScanNet~\cite{dai2017scannet}, and evaluate its performance on the held-out test splits of both. For RealEstate10K, we adopt the conventional partition of 67,477 training scenes and 7,289 test scenes. As for ScanNet, which consists of 1,513 videos of indoor scenes, we follow past work~\cite{zhang2022nerfusion, gao2023surfelnerf, wang2024freesplat} in using roughly 100 scenes for training and 8 scenes for evaluation. These datasets span a wide variety of environments, including indoor and outdoor real-estate walkthroughs (RealEstate10K), and real-world videos of numerous scenes suitable for indoor robot applications (ScanNet). We resize training and test images to $256\times256$.

\noindent \textbf{Baselines.}
We benchmark VolSplat against several recent feed-forward methods for sparse-view novel view synthesis, including both pixel-aligned and enhanced pixel-aligned Gaussian splatting approaches. Pixel-aligned methods predict Gaussian parameters on a per-pixel basis in image space before unprojecting to 3D. These include pixelSplat~\cite{charatan2024pixelsplat}, MVSplat~\cite{chen2024mvsplat}, FreeSplat~\cite{wang2024freesplat}, TranSplat~\cite{kim2025transplat} and DepthSplat~\cite{xu2025depthsplat}. Gaussian Graph Network (GGN)~\cite{zhang2024gaussian}, AnySplat~\cite{jiang2025anysplat} and WorldMirror~\cite{liu2025worldmirror} refines the pixel-aligned approach by modeling the relationships between groups of predicted Gaussians across different views while building upon it.
In contrast to both pixel-aligned and enhanced pixel-aligned methods, our VolSplat employs a voxel-aligned approach, predicting Gaussian primitives within a 3D voxel grid. This method aggregates multi-view evidence in 3D space, aligning Gaussian predictions to a voxel structure, which facilitates better geometric consistency and efficient redundancy reduction.

\noindent\textbf{Metrics.}
For quantitative evaluation, we adopt standard image quality metrics commonly used in NVS, including pixel-level PSNR, patch-level SSIM~\cite{wang2004image},and feature-level LPIPS~\cite{zhang2018unreasonable}.

\begin{figure}[t!]
  \centering
  \includegraphics[width=1.0\textwidth]{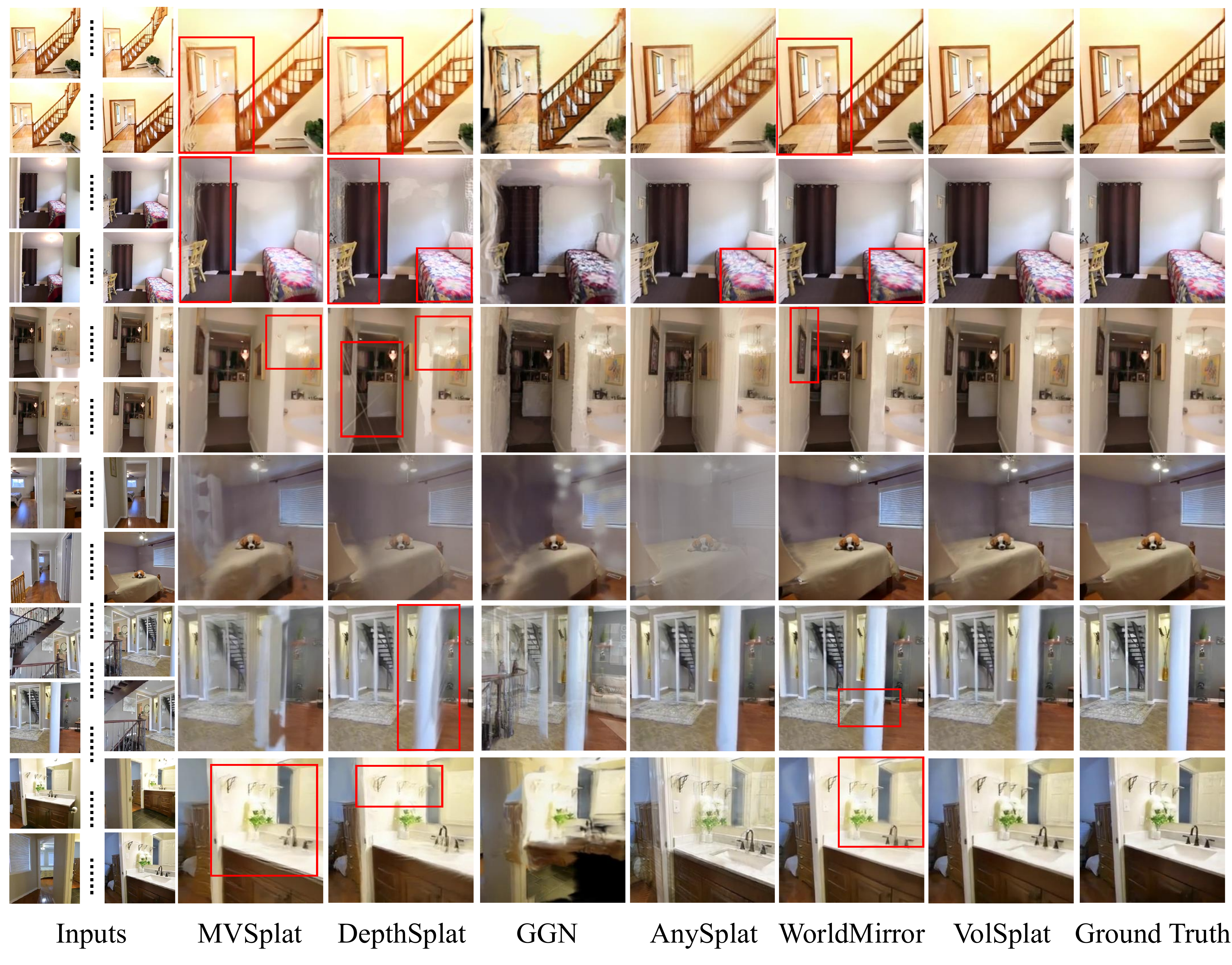}

    \vspace{-0.3cm}
  \caption{\textbf{Qualitative comparison on RealEstate10K~\cite{zhou2018stereo}.} We compare VolSplat against SOTA pixel-aligned baselines under sparse-view inputs. While competing methods often suffer from blurring and geometric distortions in complex environments, VolSplat leverages its voxel-aligned prediction to maintain superior visual fidelity and structural consistency.}
  \label{fig:vis_re10k} 
    \vspace{-0.6cm}
\end{figure}

\begin{table}[t]
  \centering
  \caption{\textbf{Quantitative comparisons on RealEstate10K~\cite{zhou2018stereo}.} The top section are pixel-aligned methods, and the middle section are methods that performs post-processing on pixel-aligned Gaussians. All baselines are retrained for fair comparison. Ground truth camera poses are provided for both AnySplat~\cite{jiang2025anysplat} and WorldMirror~\cite{liu2025worldmirror}. ``OOM'' represents that model cannot infer on a 96G GPU. We compare VolSplat against pixel-aligned and post-processing baselines under 6, 12, and 24 input views. VolSplat consistently achieves the best performance across all metrics and view settings.}
  \label{tab:re10k_multiview}
  \resizebox{\textwidth}{!}{%
  \begin{tabular}{l|ccc|ccc|ccc}
    \toprule
    \multirow{2}{*}{Method}
    & \multicolumn{3}{c|}{6v} & \multicolumn{3}{c|}{12v} & \multicolumn{3}{c}{24v} \\
    \cmidrule(lr){2-4} \cmidrule(lr){5-7} \cmidrule(lr){8-10}
    & PSNR$\uparrow$ & SSIM$\uparrow$ & LPIPS$\downarrow$ & PSNR$\uparrow$ & SSIM$\uparrow$ & LPIPS$\downarrow$ & PSNR$\uparrow$ & SSIM$\uparrow$ & LPIPS$\downarrow$ \\
    \midrule
    pixelSplat \cite{charatan2024pixelsplat} & 28.95 & 0.900 & 0.163 & OOM & OOM & OOM & OOM & OOM & OOM \\
    MVSplat \cite{chen2024mvsplat}           & 29.13 & 0.924 & 0.091 & 26.97 & 0.912 & 0.101 & 26.23 & 0.903 & 0.108 \\
    TranSplat \cite{kim2025transplat}        & 29.62 & 0.928 & 0.084 & 28.00 & 0.920 & 0.089 & 26.65 & 0.884 & 0.115 \\
    DepthSplat \cite{xu2025depthsplat}       & 30.52 & 0.931 & 0.079 & 28.54 & 0.919 & 0.088 & 26.26 & 0.880 & 0.115 \\
    \midrule
    GGN~\cite{zhang2024gaussian}    & 26.68 & 0.879 & 0.133 & 25.83 & 0.870 & 0.142 & 23.86 & 0.840 & 0.171 \\
    AnySplat~\cite{jiang2025anysplat}                                 & 19.05 & 0.576 & 0.305 & 19.86 & 0.627 & 0.288 & 20.21 & 0.658 & 0.279 \\
    WorldMirror~\cite{liu2025worldmirror}          & 24.86 & 0.819 & 0.079 & 26.03 & 0.859 & 0.072 & 26.47 & 0.875 & 0.071 \\
    \midrule
    \textbf{Ours}                            & \textbf{31.30} & \textbf{0.941} & \textbf{0.075} & \textbf{29.40} & \textbf{0.928} & \textbf{0.085} & \textbf{27.21} & \textbf{0.896} & \textbf{0.111} \\
    \bottomrule
  \end{tabular}
  }
  \vspace{-0.3cm}
\end{table}

\begin{figure}[t!]
  \centering
  \includegraphics[width=1.0\textwidth]{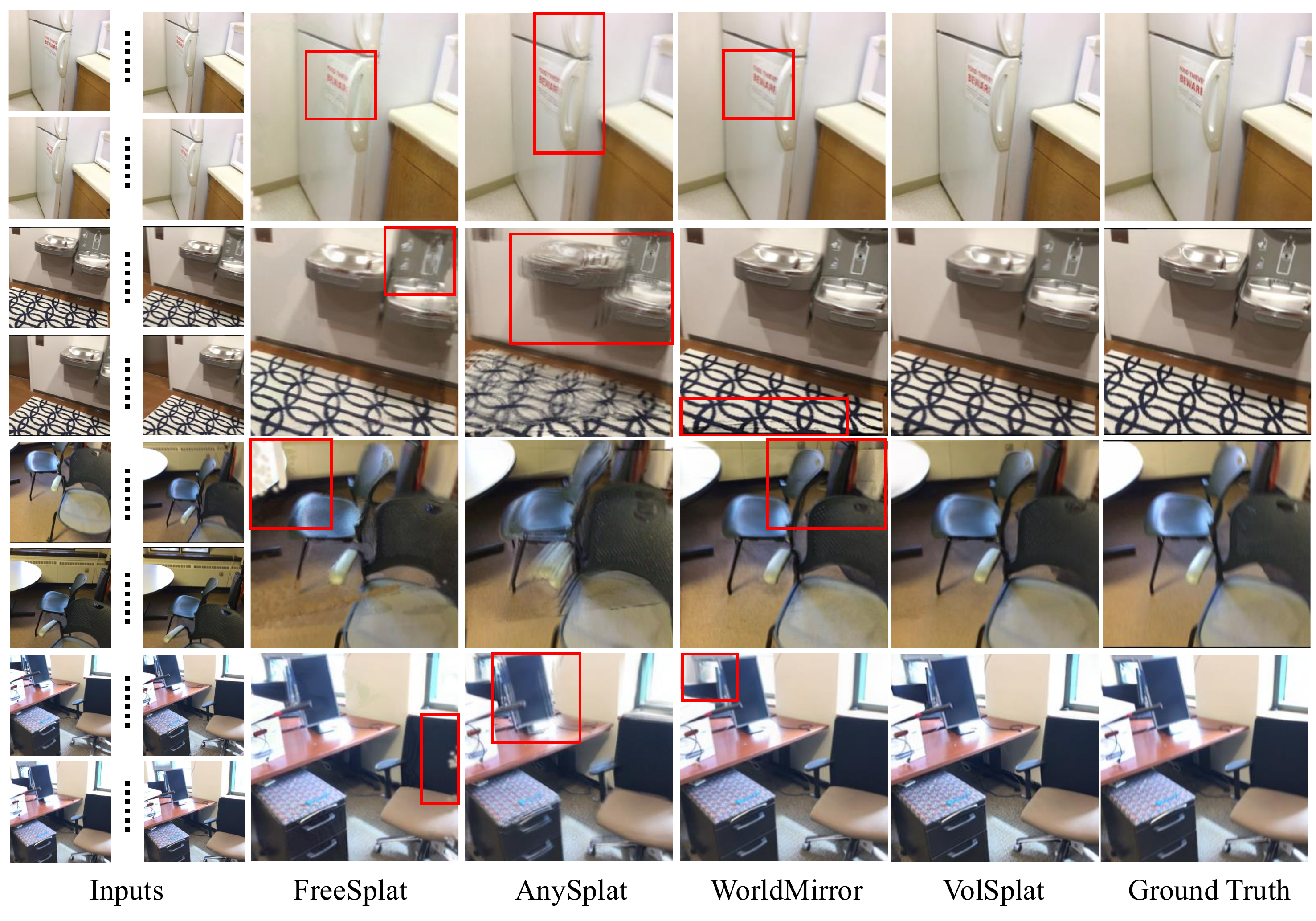}

    \vspace{-0.3cm}
  \caption{\textbf{Qualitative comparison on ScanNet~\cite{dai2017scannet}.} Compared to recent baseline methods (FreeSplat~\cite{wang2024freesplat}, AnySplat~\cite{jiang2025anysplat}, and WorldMirror~\cite{liu2025worldmirror}), VolSplat significantly reduces common floaters and visual artifacts. Our method produces cleaner object boundaries and a more coherent 3D scene reconstruction.}
  \label{fig:vis_scan} 
\end{figure}

\noindent\textbf{Implementation details.}
We implement VolSplat using PyTorch~\cite{paszke2019pytorch} and optimize the model with the AdamW~\cite{loshchilov2017decoupled} optimizer and a cosine learning rate schedule. The monocular Vision Transformer backbone is implemented using the xFormers~\cite{xFormers2022} library. For the pre-trained Depth Anything V2~\cite{depth_anything_v2} backbone, we use a lower learning rate of $2 \times 10^{-6}$, while other layers are trained with a learning rate of $2 \times 10^{-4}$ following DepthSplat~\cite{xu2025depthsplat}.
For experiments on the RealEstate10K~\cite{zhou2018stereo} and ScanNet~\cite{dai2017scannet} dataset, we train the model for 150,000 iterations using $4\times$ NVIDIA H20 GPUs with a total batch size of 4. Following the
setting of the baseline, we use $256\times256$ as input resolution. In the training stage, the number of input views is set to 6, and we evaluate the model's performance with same numbers of input views. We will make our codes and pre-trained models publicly available.

\begin{figure}[t!]
  \centering
  \includegraphics[width=1.0\textwidth]{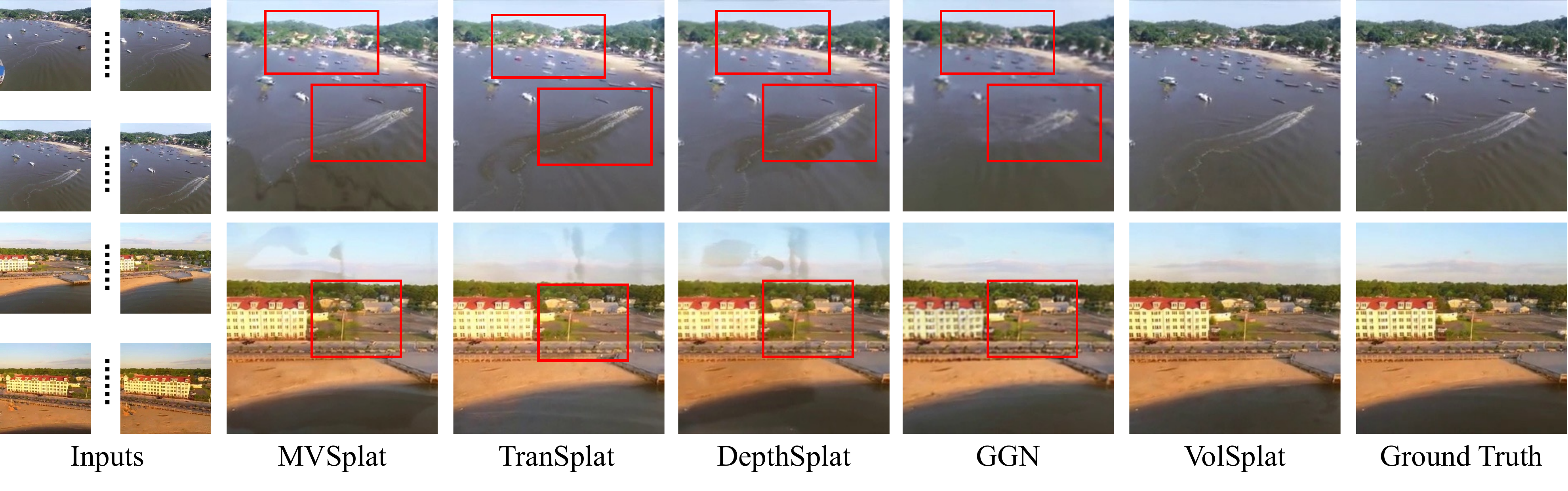}
    
    \vspace{-0.3cm}
  \caption{\textbf{Qualitative results on ACID~\cite{liu2021infinite}.} Despite being trained solely on RealEstate10K~\cite{zhou2018stereo}, VolSplat generalizes exceptionally well to outdoor scenes. Our method reconstructs fine-grained details in natural landscapes and structures, producing photorealistic novel views where baseline methods often fail to maintain visual coherence.}
  \label{fig:vis_acid} 
    \vspace{-0.3cm}
\end{figure}

\begin{figure}[!t]
  \centering
  \includegraphics[width=1.0\linewidth]{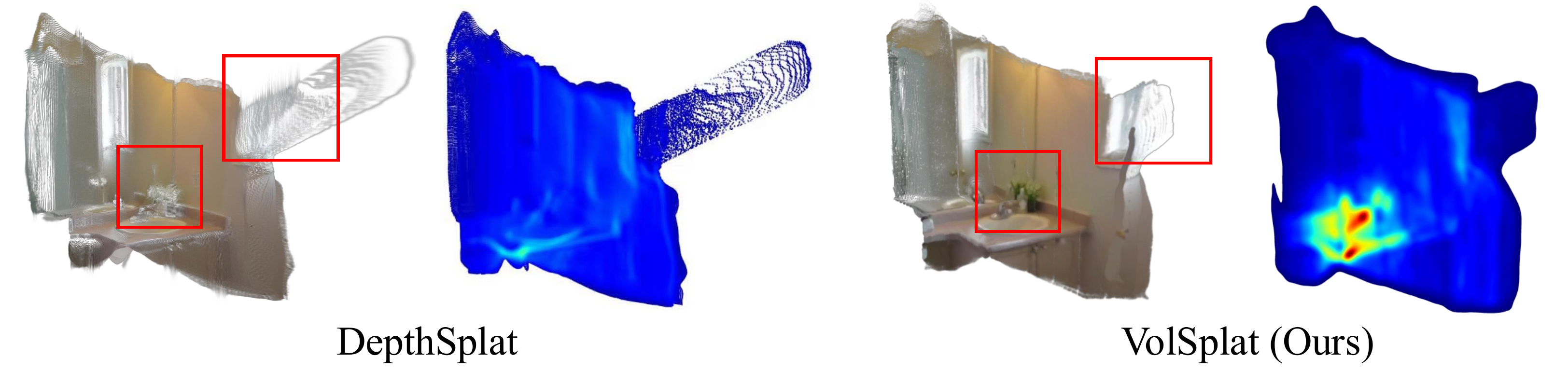}
  \caption{\textbf{Visualization of Gaussians and density maps.} We compare the rendered results and the spatial distribution of Gaussian centers between DepthSplat~\cite{xu2025depthsplat} and VolSplat. DepthSplat is constrained by the pixel grid, resulting in a uniform but redundant distribution regardless of scene content. In contrast, VolSplat adaptively concentrates Gaussians on complex geometric structures (e.g., the washbasin boundaries) while remaining sparse in flat or empty regions, demonstrating a much more efficient and geometry-aware representation.}
  \label{fig:pointmap} 
\end{figure}

\begin{table}[t]
  \centering
  \begin{minipage}[t]{0.48\linewidth}
    \centering
    \caption{\textbf{Quantitative comparisons on ScanNet~\cite{dai2017scannet} (6 input views).} We compare VolSplat against SOTA methods that rely on post-hoc 3D Gaussian fusion strategies. VolSplat outperforms all baselines. 
    }
    \label{tab:scannet_6view}
    \resizebox{\linewidth}{!}{
      \begin{tabular}{lccc}
        \toprule
        Method & {PSNR $\uparrow$} & {SSIM $\uparrow$} & {LPIPS $\downarrow$} \\
        \midrule
        FreeSplat \cite{wang2024freesplat}      & 27.45 & 0.829 & 0.222 \\
        FreeSplat++ \cite{wang2025freesplat}    & 27.45 & 0.829 & 0.223 \\
        AnySplat \cite{jiang2025anysplat}             & 19.45 & 0.626 & 0.344 \\
        WorldMirror \cite{liu2025worldmirror}      & 25.83 & 0.819 & 0.136 \\
        \midrule
        \textbf{VolSplat}  & \textbf{28.41} & \textbf{0.906} & \textbf{0.127} \\
        \bottomrule
      \end{tabular}
    }
  \end{minipage}
  \hfill %
  \begin{minipage}[t]{0.48\linewidth}
    \centering
    \caption{\textbf{Cross-dataset generalization on ACID~\cite{liu2021infinite} (6 input views).} Models trained on RealEstate10K~\cite{zhou2018stereo} (indoor scenes) are directly evaluated on ACID~\cite{liu2021infinite} (outdoor scenes) without fine-tuning. 
    }
    \label{tab:cross}
    \resizebox{\linewidth}{!}{
      \begin{tabular}{lccc}
        \toprule
        Method & {PSNR $\uparrow$} & {SSIM $\uparrow$} & {LPIPS $\downarrow$} \\
        \midrule
        MVSplat \cite{chen2024mvsplat}       & 28.15 & 0.841 & 0.147 \\
        TranSplat \cite{kim2025transplat}        & 28.17 & 0.842 & 0.146 \\
        DepthSplat \cite{xu2025depthsplat}           & 28.37 & 0.847 & 0.141 \\
        GGN \cite{zhang2024gaussian}           & 26.97 & 0.814 & 0.196 \\
        \midrule
        \textbf{VolSplat}  & \textbf{32.65} & \textbf{0.932} & \textbf{0.092} \\
        \bottomrule
      \end{tabular}
    }
  \end{minipage}
  \vspace{-0.5cm}
\end{table}

\subsection{Experimental Results and Analysis}

\noindent\textbf{Comparisons with SOTA models.}
As shown in ~\cref{tab:re10k_multiview} and ~\cref{tab:scannet_6view}, we report VolSplat's performance compared to current mainstream pixel-aligned models~\cite{charatan2024pixelsplat,chen2024mvsplat,kim2025transplat,wang2024freesplat,xu2025depthsplat} and their variants~\cite{zhang2024gaussian,jiang2025anysplat,liu2025worldmirror}. On both the RealEstate10K~\cite{zhou2018stereo} and ScanNet~\cite{dai2017scannet} datasets, VolSplat achieves SOTA results. Notably, since AnySplat~\cite{jiang2025anysplat} and WorldMirror~\cite{liu2025worldmirror} are primarily designed for joint pose-geometry optimization in pose-free settings, they fail to effectively utilize the provided ground truth camera poses, resulting in lower performance compared to other baselines.
Our experiments reveal a critical distinction between pixel-aligned and voxel-aligned paradigms. A key observation is that under sparse multi-view settings, all pixel-aligned models exhibit a significant degradation in performance. In contrast, VolSplat demonstrates promising performance to these challenging conditions. As illustrated in ~\cref{fig:vis_re10k} and ~\cref{fig:vis_scan}, images rendered by our method are largely free of the common floaters and artifacts that plague competing methods at object boundaries. This visual improvement stems directly from the ability of our model to resolve multi-view alignment issues within its 3D feature representation, resulting in cleaner edges and a more coherent 3D scene reconstruction.

\noindent\textbf{Cross-dataset generalization.}
We assess the generalization capabilities of our model on unseen outdoor datasets to verify its broad reliability. To this end, we conducted a cross-dataset generalization experiment by taking our model pre-trained on the RealEstate10K~\cite{zhou2018stereo} dataset and evaluating it directly on the ACID~\cite{liu2021infinite} dataset without any fine-tuning.
As demonstrated in~\cref{tab:cross} and \cref{fig:vis_acid}, VolSplat maintains significantly higher performance in this zero-shot transfer setting. We attribute this superior generalization to the inherent robustness of our voxel-aligned framework. Pixel-aligned models exhibit a much higher sensitivity to the variations in data complexity and distribution between different datasets. In contrast, VolSplat is less susceptible to these domain shifts.

\noindent\textbf{Analysis of Gaussian density.}
A fundamental principle of 3D reconstruction is that the complexity of the representation should adapt to the complexity of the scene. Real-world environments contain a mix of simple, planar surfaces and intricate, high-frequency geometric details. An ideal model should allocate its descriptive capacity accordingly. However, pixel-aligned methods are inherently limited in this regard. Their paradigm of predicting one Gaussian per pixel results in a fixed number of primitives, predetermined by the input image resolution (e.g., $H\times W$ Gaussians from a reference view), regardless of whether the scene is a simple room or a complex outdoor environment.
\begin{table}[t]
  \centering
  \caption{\textbf{Analysis of voxel size.} ``PGS'' stands for ``average number of per-view Gaussians''. We investigate the impact of voxel resolution on reconstruction quality and efficiency. A voxel size of 0.1 yields the optimal trade-off, achieving the best performance. Notably, further reducing the voxel size to 0.05 degrades performance due to the loss of coherent spatial context in overly sparse grids, while larger voxels fail to capture fine geometric details.}
  \label{tab:voxel}
  \begin{tabular}{lcccccc}
    \toprule
    {Voxel Size (cm)} & {PSNR $\uparrow$} & {SSIM $\uparrow$} & {LPIPS $\downarrow$} & {PGS} & Memory(GB) & Inference Time(s)\\
    \midrule
    0.05       & 29.34 & 0.919 & 0.092 & 65415 & 9.19 & 0.802 \\
    0.1 (default)    & \textbf{29.40} & \textbf{0.928} & \textbf{0.085} & 60523 & 9.04 & 0.768 \\
    0.5       & 27.33 & 0.899 & 0.108 & 59788 & 8.98 & 0.744 \\
    1       & 20.78 & 0.602 & 0.323 & 51806 & 8.74 & 0.739 \\
    \bottomrule
  \end{tabular}
\end{table}

\begin{table}[t]
  \centering
  \caption{\textbf{Ablation of sparse 3D decoder.} ``w/ 3D CNN'' means replacing the 3D U-Net with a sparse 3D CNN, ``w/o residual'' means predicting refined voxel feature without residual design, and ``w/o decoder'' means removing the refinement stage. We validate the necessity of our specific refinement module. Replacing our proposed sparse 3D decoder for feature refinement results in performance drop. This demonstrates that our specific design is essential for refining the voxel features and producing high-quality 3D Gaussians.}
  \label{tab:ablation}
  \begin{tabular}{lccccc}
    \toprule
    Components & {PSNR $\uparrow$} & {SSIM $\uparrow$} & {LPIPS $\downarrow$} & Memory(GB) & Inference Time(s)\\
    \midrule
    default     & \textbf{29.40} & \textbf{0.928} & \textbf{0.085} & 9.04 & 0.768 \\
    w/ 3D CNN   & 28.01 & 0.919 & 0.098 & 9.03 & 0.705 \\
    w/o residual   & 27.92 & 0.908 & 0.101 & 9.04 & 0.765 \\
    w/o decoder   & 27.47 & 0.901 & 0.102 & 8.99 & 0.687 \\
    \bottomrule
  \end{tabular}
  \vspace{-0.3cm}
\end{table}
In stark contrast, our voxel-aligned framework enables adaptive control over the density of the 3D Gaussians. 
By predicting primitives based on the occupancy of 3D voxel features, VolSplat naturally allocates a higher concentration of Gaussians to regions of high geometric detail while using a sparser representation for simple or empty spaces.

This adaptive capability is quantitatively validated by the results we reported in~\cref{tab:re10k_multiview},~\cref{tab:scannet_6view} and~\cref{tab:cross}. Here, we analyze these findings in greater detail. The data shows that pixel-aligned methods consistently generate constant density of Gaussians, irrespective of the scene content. This leads to significant redundancy, as well as an insufficient representational capacity in areas with intricate details. Conversely, Gaussians of VolSplat demonstrate significant variance across different regions, confirming its ability to tailor complexity of the scene, as shown in ~\cref{fig:pointmap}. Notably, VolSplat often achieves superior rendering quality with a non-uniform set of Gaussians compared to the brute-force density of pixel-aligned approaches.

\subsection{Ablation Study}

In this section, we study the properties of our key components with 12 input views on the RealEstate10K~\cite{zhou2018stereo} dataset.

\noindent \textbf{Ablation of Voxel Size.}
The voxel size is a critical hyperparameter in our framework, as it dictates the resolution of the 3D feature grid. This choice involves a fundamental trade-off between the fidelity of the geometric representation and computational resource consumption. In~\cref{tab:voxel}, we analyze this trade-off by comparing our default setting against configurations with different voxels.

Using a small voxel size increases the granularity of the 3D grid, allowing the model to capture finer geometric details. It comes at a significant cost, substantially increasing memory usage and processing time due to the cubic growth of the voxel volume. Conversely, employing a large voxel size reduces the computational footprint but results in a coarser quantization of the 3D space. Our default configuration strikes an effective balance, achieving SOTA performance while maintaining manageable computational requirements.

\noindent\textbf{Ablation of Model Architecture.}
Directly predicting Gaussians from the initial unprojected 3D features is less effective, particularly for challenging scenes with complex geometry or sparse viewpoints. To address this, we incorporate a 3D U-Net architecture to refine and enhance this raw feature volume, predicting the residual features. To validate the necessity and efficacy of this design, we conduct an ablation study with three variants: 1) removing the refinement module entirely, 2)directly predicting the refined feature, and 3) replacing the 3D U-Net with a standard 3D CNN.

The results, presented in ~\cref{tab:ablation}, confirm our architectural choices. Removing the refinement stage altogether leads to a significant drop in performance, demonstrating that processing the initial voxel features is critical for producing a coherent 3D representation. While substituting our module with a sparse 3D CNN or removing the residual design yields better results than no refinement, it still falls short of the performance of our full model. The multi-scale feature fusion inherent in the U-Net structure and residual design are crucial for capturing both fine-grained local details and broader spatial context. 

\section{Conclusion}

We address the fundamental limitations inherent in the prevailing pixel-aligned paradigm for feed-forward 3D Gaussian Splatting. We identify that existing methods suffer from a rigid coupling of Gaussian density to input image resolution and a high sensitivity to multi-view alignment errors. To overcome these challenges, we introduce VolSplat, a novel framework that fundamentally shifts the reconstruction process from 2D pixels to a 3D voxel-aligned space. By constructing 3D voxel feature and predicting Gaussians directly from this unified representation, our method effectively decouples the 3D scene from the constraints of the input views.
This voxel-centric design enables adaptive control over Gaussian density according to scene complexity and inherently resolves alignment ambiguities, leading to more geometrically consistent and faithful reconstructions for downstream tasks.

\bibliographystyle{unsrtnat}
\bibliography{main}

@String(PAMI  = {IEEE Trans. Pattern Anal. Mach. Intell.})

@String(CVPR  = {IEEE Conf. Comput. Vis. Pattern Recog.})

@String(ICCV  = {Int. Conf. Comput. Vis.})

@String(ECCV  = {Eur. Conf. Comput. Vis.})

@String(NeurIPS = {Adv. Neural Inform. Process. Syst.})

@String(AAAI  = {AAAI})

@String(TOG   = {ACM Trans. Graph.})

@String(TIP   = {IEEE Trans. Image Process.})

@String(ICRA  = {IEEE Int. Conf. Robot. Autom.})

@article{wang2026latent,
  title={Latent Spatial Memory for Video World Models},
  author={Wang, Weijie and Zhao, Haoyu and Yang, Yifan and Chen, Feng and Zhang, Zeyu and He, Yefei and Duan, Zicheng and Chen, Donny Y and Yang, Yuqing and Zhuang, Bohan},
  journal={arXiv preprint arXiv:2606.09828},
  year={2026}
}

@inproceedings{barron2023zip,
  title={Zip-nerf: Anti-aliased grid-based neural radiance fields},
  author={Barron, Jonathan T and Mildenhall, Ben and Verbin, Dor and Srinivasan, Pratul P and Hedman, Peter},
  booktitle=ICCV,
  pages={19697--19705},
  year={2023}
}

@inproceedings{barron2021mip,
  title={Mip-nerf: A multiscale representation for anti-aliasing neural radiance fields},
  author={Barron, Jonathan T and Mildenhall, Ben and Tancik, Matthew and Hedman, Peter and Martin-Brualla, Ricardo and Srinivasan, Pratul P},
  booktitle=ICCV,
  pages={5855--5864},
  year={2021}
}

@article{mildenhall2021nerf,
  title={Nerf: Representing scenes as neural radiance fields for view synthesis},
  author={Mildenhall, Ben and Srinivasan, Pratul P and Tancik, Matthew and Barron, Jonathan T and Ramamoorthi, Ravi and Ng, Ren},
  journal={Communications of the ACM},
  volume={65},
  number={1},
  pages={99--106},
  year={2021},
  publisher={ACM New York, NY, USA}
}

@article{kerbl20233d,
  title={3D Gaussian splatting for real-time radiance field rendering.},
  author={Kerbl, Bernhard and Kopanas, Georgios and Leimk{\"u}hler, Thomas and Drettakis, George},
  journal=TOG,
  volume={42},
  number={4},
  pages={139--1},
  year={2023}
}

@article{meagher1982geometric,
  title={Geometric modeling using octree encoding},
  author={Meagher, Donald},
  journal={Computer graphics and image processing},
  volume={19},
  number={2},
  pages={129--147},
  year={1982},
  publisher={Elsevier}
}

@article{kaufman2005overview,
  title={Overview of volume rendering.},
  author={Kaufman, Arie E and Mueller, Klaus},
  journal={The visualization handbook},
  volume={7},
  pages={127--174},
  year={2005}
}

@inproceedings{koneputugodage2023octree,
  title={Octree guided unoriented surface reconstruction},
  author={Koneputugodage, Chamin Hewa and Ben-Shabat, Yizhak and Gould, Stephen},
  booktitle=CVPR,
  pages={16717--16726},
  year={2023}
}

@inproceedings{zhou2018voxelnet,
  title={Voxelnet: End-to-end learning for point cloud based 3d object detection},
  author={Zhou, Yin and Tuzel, Oncel},
  booktitle=CVPR,
  pages={4490--4499},
  year={2018}
}

@inproceedings{riegler2017octnet,
  title={Octnet: Learning deep 3d representations at high resolutions},
  author={Riegler, Gernot and Osman Ulusoy, Ali and Geiger, Andreas},
  booktitle=CVPR,
  pages={3577--3586},
  year={2017}
}

@inproceedings{fridovich2022plenoxels,
  title={Plenoxels: Radiance fields without neural networks},
  author={Fridovich-Keil, Sara and Yu, Alex and Tancik, Matthew and Chen, Qinhong and Recht, Benjamin and Kanazawa, Angjoo},
  booktitle=CVPR,
  pages={5501--5510},
  year={2022}
}

@inproceedings{fridovich2023k,
  title={K-planes: Explicit radiance fields in space, time, and appearance},
  author={Fridovich-Keil, Sara and Meanti, Giacomo and Warburg, Frederik Rahb{\ae}k and Recht, Benjamin and Kanazawa, Angjoo},
  booktitle=CVPR,
  pages={12479--12488},
  year={2023}
}

@inproceedings{lu2024scaffold,
  title={Scaffold-gs: Structured 3d gaussians for view-adaptive rendering},
  author={Lu, Tao and Yu, Mulin and Xu, Linning and Xiangli, Yuanbo and Wang, Limin and Lin, Dahua and Dai, Bo},
  booktitle=CVPR,
  pages={20654--20664},
  year={2024}
}

@article{fan2024lightgaussian,
  title={Lightgaussian: Unbounded 3d gaussian compression with 15x reduction and 200+ fps},
  author={Fan, Zhiwen and Wang, Kevin and Wen, Kairun and Zhu, Zehao and Xu, Dejia and Wang, Zhangyang and others},
  journal=NeurIPS,
  volume={37},
  pages={140138--140158},
  year={2024}
}

@inproceedings{liu2025flexgs,
  title={FlexGS: Train Once, Deploy Everywhere with Many-in-One Flexible 3D Gaussian Splatting},
  author={Liu, Hengyu and Wang, Yuehao and Li, Chenxin and Cai, Ruisi and Wang, Kevin and Li, Wuyang and Molchanov, Pavlo and Wang, Peihao and Wang, Zhangyang},
  booktitle=CVPR,
  pages={16336--16345},
  year={2025}
}

@article{zhang2024gaussian,
  title={Gaussian graph network: Learning efficient and generalizable gaussian representations from multi-view images},
  author={Zhang, Shengjun and Fei, Xin and Liu, Fangfu and Song, Haixu and Duan, Yueqi},
  journal=NeurIPS,
  volume={37},
  pages={50361--50380},
  year={2024}
}

@article{ziwen2024long,
  title={Long-lrm: Long-sequence large reconstruction model for wide-coverage gaussian splats},
  author={Ziwen, Chen and Tan, Hao and Zhang, Kai and Bi, Sai and Luan, Fujun and Hong, Yicong and Fuxin, Li and Xu, Zexiang},
  journal={arXiv preprint arXiv:2410.12781},
  year={2024}
}

@inproceedings{ji2017deep,
  title={Deep view morphing},
  author={Ji, Dinghuang and Kwon, Junghyun and McFarland, Max and Savarese, Silvio},
  booktitle=CVPR,
  pages={2155--2163},
  year={2017}
}

@inproceedings{levoy2023light,
author = {Levoy, Marc and Hanrahan, Pat},
title = {Light field rendering},
year = {1996},
isbn = {0897917464},
publisher = {Association for Computing Machinery},
address = {New York, NY, USA},
doi = {10.1145/237170.237199},
booktitle = {Proceedings of the 23rd Annual Conference on Computer Graphics and Interactive Techniques},
pages = {31-42},
numpages = {12},
series = {SIGGRAPH '96}
}

@inproceedings{debevec2023modeling,
author = {Debevec, Paul E. and Taylor, Camillo J. and Malik, Jitendra},
title = {Modeling and rendering architecture from photographs: a hybrid geometry- and image-based approach},
year = {1996},
isbn = {0897917464},
publisher = {Association for Computing Machinery},
address = {New York, NY, USA},
doi = {10.1145/237170.237191},
booktitle = {Proceedings of the 23rd Annual Conference on Computer Graphics and Interactive Techniques},
pages = {11-20},
numpages = {10},
series = {SIGGRAPH '96}
}

@inproceedings{zhu2024fsgs,
  title={Fsgs: Real-time few-shot view synthesis using gaussian splatting},
  author={Zhu, Zehao and Fan, Zhiwen and Jiang, Yifan and Wang, Zhangyang},
  booktitle=ECCV,
  pages={145--163},
  year={2024},
  organization={Springer}
}

@article{ren2024octree,
  title={Octree-gs: Towards consistent real-time rendering with lod-structured 3d gaussians},
  author={Ren, Kerui and Jiang, Lihan and Lu, Tao and Yu, Mulin and Xu, Linning and Ni, Zhangkai and Dai, Bo},
  journal={arXiv preprint arXiv:2403.17898},
  year={2024}
}

@inproceedings{charatan2024pixelsplat,
  title={pixelsplat: 3d gaussian splats from image pairs for scalable generalizable 3d reconstruction},
  author={Charatan, David and Li, Sizhe Lester and Tagliasacchi, Andrea and Sitzmann, Vincent},
  booktitle=CVPR,
  pages={19457--19467},
  year={2024}
}

@inproceedings{chen2024mvsplat,
  title={Mvsplat: Efficient 3d gaussian splatting from sparse multi-view images},
  author={Chen, Yuedong and Xu, Haofei and Zheng, Chuanxia and Zhuang, Bohan and Pollefeys, Marc and Geiger, Andreas and Cham, Tat-Jen and Cai, Jianfei},
  booktitle=ECCV,
  pages={370--386},
  year={2024},
  organization={Springer}
}

@inproceedings{xu2025depthsplat,
  title={Depthsplat: Connecting gaussian splatting and depth},
  author={Xu, Haofei and Peng, Songyou and Wang, Fangjinhua and Blum, Hermann and Barath, Daniel and Geiger, Andreas and Pollefeys, Marc},
  booktitle=CVPR,
  pages={16453--16463},
  year={2025}
}

@article{wang2024freesplat,
  title={Freesplat: Generalizable 3d gaussian splatting towards free view synthesis of indoor scenes},
  author={Wang, Yunsong and Huang, Tianxin and Chen, Hanlin and Lee, Gim Hee},
  journal=NeurIPS,
  volume={37},
  pages={107326--107349},
  year={2024}
}

@article{wang2025zpressor,
  title={ZPressor: Bottleneck-Aware Compression for Scalable Feed-Forward 3DGS},
  author={Wang, Weijie and Chen, Donny Y and Zhang, Zeyu and Shi, Duochao and Liu, Akide and Zhuang, Bohan},
  journal={arXiv preprint arXiv:2505.23734},
  year={2025}
}

@article{chen2024mvsplat360,
  title={Mvsplat360: Feed-forward 360 scene synthesis from sparse views},
  author={Chen, Yuedong and Zheng, Chuanxia and Xu, Haofei and Zhuang, Bohan and Vedaldi, Andrea and Cham, Tat-Jen and Cai, Jianfei},
  journal=NeurIPS,
  volume={37},
  pages={107064--107086},
  year={2024}
}

@article{ye2024no,
  title={No pose, no problem: Surprisingly simple 3d gaussian splats from sparse unposed images},
  author={Ye, Botao and Liu, Sifei and Xu, Haofei and Li, Xueting and Pollefeys, Marc and Yang, Ming-Hsuan and Peng, Songyou},
  journal={arXiv preprint arXiv:2410.24207},
  year={2024}
}

@inproceedings{kang2025selfsplat,
  title={SelfSplat: Pose-free and 3D prior-free generalizable 3D Gaussian splatting},
  author={Kang, Gyeongjin and Yoo, Jisang and Park, Jihyeon and Nam, Seungtae and Im, Hyeonsoo and Shin, Sangheon and Kim, Sangpil and Park, Eunbyung},
  booktitle=CVPR,
  pages={22012--22022},
  year={2025}
}

@inproceedings{he2016deep,
  title={Deep residual learning for image recognition},
  author={He, Kaiming and Zhang, Xiangyu and Ren, Shaoqing and Sun, Jian},
  booktitle=CVPR,
  pages={770--778},
  year={2016}
}

@inproceedings{liu2021swin,
  title={Swin transformer: Hierarchical vision transformer using shifted windows},
  author={Liu, Ze and Lin, Yutong and Cao, Yue and Hu, Han and Wei, Yixuan and Zhang, Zheng and Lin, Stephen and Guo, Baining},
  booktitle=ICCV,
  pages={10012--10022},
  year={2021}
}

@article{xu2023unifying,
  title={Unifying flow, stereo and depth estimation},
  author={Xu, Haofei and Zhang, Jing and Cai, Jianfei and Rezatofighi, Hamid and Yu, Fisher and Tao, Dacheng and Geiger, Andreas},
  journal=PAMI,
  volume={45},
  number={11},
  pages={13941--13958},
  year={2023},
  publisher={IEEE}
}

@inproceedings{wang2024embodiedscan,
  title={Embodiedscan: A holistic multi-modal 3d perception suite towards embodied ai},
  author={Wang, Tai and Mao, Xiaohan and Zhu, Chenming and Xu, Runsen and Lyu, Ruiyuan and Li, Peisen and Chen, Xiao and Zhang, Wenwei and Chen, Kai and Xue, Tianfan and others},
  booktitle=CVPR,
  pages={19757--19767},
  year={2024}
}

@inproceedings{zhang2018unreasonable,
  title={The unreasonable effectiveness of deep features as a perceptual metric},
  author={Zhang, Richard and Isola, Phillip and Efros, Alexei A and Shechtman, Eli and Wang, Oliver},
  booktitle=CVPR,
  pages={586--595},
  year={2018}
}

@article{wang2026world,
  title={World-r1: Reinforcing 3d constraints for text-to-video generation},
  author={Wang, Weijie and He, Xiaoxuan and Gu, Youping and Yang, Yifan and Zhang, Zeyu and He, Yefei and Ding, Yanbo and Hu, Xirui and Chen, Donny Y and He, Zhiyuan and others},
  journal={arXiv preprint arXiv:2604.24764},
  year={2026}
}

@inproceedings{zhang2026panflow,
  title={Panflow: Decoupled motion control for panoramic video generation},
  author={Zhang, Cheng and Liang, Hanwen and Chen, Donny Y and Wu, Qianyi and Plataniotis, Konstantinos N and Gambardella, Camilo Cruz and Cai, Jianfei},
  booktitle=AAAI,
  volume={40},
  number={15},
  pages={12385--12393},
  year={2026}
}

@inproceedings{ni2025wonderturbo,
  title={Wonderturbo: Generating interactive 3d world in 0.72 seconds},
  author={Ni, Chaojun and Wang, Xiaofeng and Zhu, Zheng and Wang, Weijie and Li, Haoyun and Zhao, Guosheng and Li, Jie and Qin, Wenkang and Huang, Guan and Mei, Wenjun},
  booktitle=ICCV,
  pages={27423--27434},
  year={2025}
}

@article{wang2026trisplat,
  title={TriSplat: Simulation-Ready Feed-Forward 3D Scene Reconstruction},
  author={Wang, Weijie and Li, Zimu and Shi, Jinchuan and Zhang, Zeyu and Ye, Botao and Pollefeys, Marc and Chen, Donny Y and Zhuang, Bohan},
  journal={arXiv preprint arXiv:2605.26115},
  year={2026}
}

@article{wang2026feed,
  title={Feed-forward 3d scene modeling: A problem-driven perspective},
  author={Wang, Weijie and Cao, Qihang and Gao, Sensen and Chen, Donny Y and Xu, Haofei and Bian, Wenjing and Peng, Songyou and Cham, Tat-Jen and Zheng, Chuanxia and Geiger, Andreas and others},
  journal={arXiv preprint arXiv:2604.14025},
  year={2026}
}

@article{wang2025drivegen3d,
  title={Drivegen3d: Boosting feed-forward driving scene generation with efficient video diffusion},
  author={Wang, Weijie and Zhu, Jiagang and Zhang, Zeyu and Wang, Xiaofeng and Zhu, Zheng and Zhao, Guosheng and Ni, Chaojun and Wang, Haoxiao and Huang, Guan and Chen, Xinze and others},
  journal={arXiv preprint arXiv:2510.15264},
  year={2025}
}

@article{zhou2018stereo,
  title={Stereo magnification: Learning view synthesis using multiplane images},
  author={Zhou, Tinghui and Tucker, Richard and Flynn, John and Fyffe, Graham and Snavely, Noah},
  journal={arXiv preprint arXiv:1805.09817},
  year={2018}
}

@article{kang2025ilrm,
  title={iLRM: An Iterative Large 3D Reconstruction Model},
  author={Kang, Gyeongjin and Nam, Seungtae and Sun, Xiangyu and Khamis, Sameh and Mohamed, Abdelrahman and Park, Eunbyung},
  journal={arXiv preprint arXiv:2507.23277},
  year={2025}
}

@article{jiang2025anysplat,
  title={AnySplat: Feed-forward 3D Gaussian Splatting from Unconstrained Views},
  author={Jiang, Lihan and Mao, Yucheng and Xu, Linning and Lu, Tao and Ren, Kerui and Jin, Yichen and Xu, Xudong and Yu, Mulin and Pang, Jiangmiao and Zhao, Feng and Lin, Dahua and Dai, Bo},
  journal={arXiv preprint arXiv:2505.23716},
  year={2025}
}

@inproceedings{kim2025transplat,
  title={Transplat: Generalizable 3d gaussian splatting from sparse multi-view images with transformers},
  author={Zhang, Chuanrui and Zou, Yingshuang and Li, Zhuoling and Yi, Minmin and Wang, Haoqian},
  booktitle=AAAI,
  pages={9869--9877},
  year={2025}
}

@inproceedings{yu2021pixelnerf,
  title={pixelnerf: Neural radiance fields from one or few images},
  author={Yu, Alex and Ye, Vickie and Tancik, Matthew and Kanazawa, Angjoo},
  booktitle=CVPR,
  pages={4578--4587},
  year={2021}
}

@inproceedings{wang2021ibrnet,
  title={Ibrnet: Learning multi-view image-based rendering},
  author={Wang, Qianqian and Wang, Zhicheng and Genova, Kyle and Srinivasan, Pratul P and Zhou, Howard and Barron, Jonathan T and Martin-Brualla, Ricardo and Snavely, Noah and Funkhouser, Thomas},
  booktitle=CVPR,
  pages={4690--4699},
  year={2021}
}

@article{shi2025pmloss,
  title={Revisiting Depth Representations for Feed-Forward 3D Gaussian Splatting},
  author={Shi, Duochao and Wang, Weijie and Chen, Donny Y. and Zhang, Zeyu and Bian, Jiawang and Zhuang, Bohan and Shen, Chunhua},
  journal={arXiv preprint arXiv:2506.05327},
  year={2025}
}

@article{depth_anything_v2,
  title={Depth Anything V2},
  author={Yang, Lihe and Kang, Bingyi and Huang, Zilong and Zhao, Zhen and Xu, Xiaogang and Feng, Jiashi and Zhao, Hengshuang},
  journal={arXiv:2406.09414},
  year={2024}
}

@inproceedings{cciccek20163d,
  title={3D U-Net: learning dense volumetric segmentation from sparse annotation},
  author={{\c{C}}i{\c{c}}ek, {\"O}zg{\"u}n and Abdulkadir, Ahmed and Lienkamp, Soeren S and Brox, Thomas and Ronneberger, Olaf},
  booktitle={International conference on medical image computing and computer-assisted intervention},
  pages={424--432},
  year={2016},
  organization={Springer}
}

@article{odena2016deconvolution,
  author = {Odena, Augustus and Dumoulin, Vincent and Olah, Chris},
  title = {Deconvolution and Checkerboard Artifacts},
  journal = {Distill},
  year = {2016},
  url = {http://distill.pub/2016/deconv-checkerboard},
  doi = {10.23915/distill.00003}
}

@inproceedings{dai2017scannet,
  title={Scannet: Richly-annotated 3d reconstructions of indoor scenes},
  author={Dai, Angela and Chang, Angel X and Savva, Manolis and Halber, Maciej and Funkhouser, Thomas and Nie{\ss}ner, Matthias},
  booktitle=CVPR,
  pages={5828--5839},
  year={2017}
}

@inproceedings{gao2023surfelnerf,
  title={Surfelnerf: Neural surfel radiance fields for online photorealistic reconstruction of indoor scenes},
  author={Gao, Yiming and Cao, Yan-Pei and Shan, Ying},
  booktitle=CVPR,
  pages={108--118},
  year={2023}
}

@inproceedings{zhang2022nerfusion,
  title={Nerfusion: Fusing radiance fields for large-scale scene reconstruction},
  author={Zhang, Xiaoshuai and Bi, Sai and Sunkavalli, Kalyan and Su, Hao and Xu, Zexiang},
  booktitle=CVPR,
  pages={5449--5458},
  year={2022}
}

@article{paszke2019pytorch,
  title={Pytorch: An imperative style, high-performance deep learning library},
  author={Paszke, Adam and Gross, Sam and Massa, Francisco and Lerer, Adam and Bradbury, James and Chanan, Gregory and Killeen, Trevor and Lin, Zeming and Gimelshein, Natalia and Antiga, Luca and others},
  journal=NeurIPS,
  volume={32},
  year={2019}
}

@article{loshchilov2017decoupled,
  title={Decoupled weight decay regularization},
  author={Loshchilov, Ilya and Hutter, Frank},
  journal={arXiv preprint arXiv:1711.05101},
  year={2017}
}

@Misc{xFormers2022,
  author =       {Benjamin Lefaudeux and Francisco Massa and Diana Liskovich and Wenhan Xiong and Vittorio Caggiano and Sean Naren and Min Xu and Jieru Hu and Marta Tintore and Susan Zhang and Patrick Labatut and Daniel Haziza and Luca Wehrstedt and Jeremy Reizenstein and Grigory Sizov},
  title =        {xFormers: A modular and hackable Transformer modelling library},
  howpublished = {\url{https://github.com/facebookresearch/xformers}},
  year =         {2022}
}

@article{wang2026diffusion,
  title={Diffusion Model as a Generalist Segmentation Learner},
  author={Wang, Haoxiao and Xiang, Antao and Sun, Haiyang and Sun, Peilin and Pan, Changhao and Chen, Yifu and Hong, Minjie and Wang, Weijie and Chen, Shuang and Chen, Yue and others},
  journal={arXiv preprint arXiv:2604.24575},
  year={2026}
}

@inproceedings{wang2025transdiff,
  title={Transdiff: Diffusion-based method for manipulating transparent objects using a single rgb-d image},
  author={Wang, Haoxiao and Zhou, Kaichen and Gu, Binrui and Feng, Zhiyuan and Wang, Weijie and Sun, Peilin and Xiao, Yicheng and Zhang, Jianhua and Dong, Hao},
  booktitle=ICRA,
  pages={7277--7283},
  year={2025},
  organization={IEEE}
}

@inproceedings{liu2021infinite,
  title={Infinite nature: Perpetual view generation of natural scenes from a single image},
  author={Liu, Andrew and Tucker, Richard and Jampani, Varun and Makadia, Ameesh and Snavely, Noah and Kanazawa, Angjoo},
  booktitle=ICCV,
  pages={14458--14467},
  year={2021}
}

@article{wang2025freesplat,
  title={FreeSplat++: Generalizable 3D Gaussian Splatting for Efficient Indoor Scene Reconstruction},
  author={Wang, Yunsong and Huang, Tianxin and Chen, Hanlin and Lee, Gim Hee},
  journal={arXiv preprint arXiv:2503.22986},
  year={2025}
}

@inproceedings{miao2025evolsplat,
  title={Evolsplat: Efficient volume-based gaussian splatting for urban view synthesis},
  author={Miao, Sheng and Huang, Jiaxin and Bai, Dongfeng and Yan, Xu and Zhou, Hongyu and Wang, Yue and Liu, Bingbing and Geiger, Andreas and Liao, Yiyi},
  booktitle=CVPR,
  pages={11286--11296},
  year={2025}
}

@inproceedings{chen2021mvsnerf,
  title={Mvsnerf: Fast generalizable radiance field reconstruction from multi-view stereo},
  author={Chen, Anpei and Xu, Zexiang and Zhao, Fuqiang and Zhang, Xiaoshuai and Xiang, Fanbo and Yu, Jingyi and Su, Hao},
  booktitle=ICCV,
  pages={14124--14133},
  year={2021}
}

@inproceedings{zhang2024gs,
  title={Gs-lrm: Large reconstruction model for 3d gaussian splatting},
  author={Zhang, Kai and Bi, Sai and Tan, Hao and Xiangli, Yuanbo and Zhao, Nanxuan and Sunkavalli, Kalyan and Xu, Zexiang},
  booktitle=ECCV,
  pages={1--19},
  year={2024},
  organization={Springer}
}

@article{liu2025worldmirror,
  title={Worldmirror: Universal 3d world reconstruction with any-prior prompting},
  author={Liu, Yifan and Min, Zhiyuan and Wang, Zhenwei and Wu, Junta and Wang, Tengfei and Yuan, Yixuan and Luo, Yawei and Guo, Chunchao},
  journal={arXiv preprint arXiv:2510.10726},
  year={2025}
}

@article{wang2004image,
  title={Image quality assessment: from error visibility to structural similarity},
  author={Wang, Zhou and Bovik, Alan C and Sheikh, Hamid R and Simoncelli, Eero P},
  journal=TIP,
  volume={13},
  number={4},
  pages={600--612},
  year={2004},
  publisher={IEEE}
}

@article{wang2025learning,
  title={Learning Efficient Fuse-and-Refine for Feed-Forward 3D Gaussian Splatting},
  author={Wang, Yiming and Chai, Lucy and Luo, Xuan and Niemeyer, Michael and Lagunas, Manuel and Lombardi, Stephen and Tang, Siyu and Sun, Tiancheng},
  journal={arXiv preprint arXiv:2503.14698},
  year={2025}
}

@article{huang2025spfsplat,
      title={No Pose at All: Self-Supervised Pose-Free 3D Gaussian Splatting from Sparse Views},
      author={Huang, Ranran and Mikolajczyk, Krystian},
      journal={arXiv preprint arXiv: 2508.01171},
      year={2025}
}

@inproceedings{choy20194d,
  title={4D Spatio-Temporal ConvNets: Minkowski Convolutional Neural Networks},
  author={Choy, Christopher and Gwak, JunYoung and Savarese, Silvio},
  booktitle=CVPR,
  pages={3075--3084},
  year={2019}
}

\clearpage
\appendix

\setcounter{table}{0}
\setcounter{figure}{0}
\setcounter{section}{0}
\renewcommand{\thetable}{\Alph{table}}
\renewcommand{\thefigure}{\Alph{figure}}
\renewcommand{\thesection}{\Alph{section}} 

\section{More Implementation Details}
\label{sec:app_implementation}

\boldstart{Network architecture.}
Our framework begins with a 2D feature extraction stage. Following DepthSplat~\cite{xu2025depthsplat}, we employ a weight-sharing ResNet~\cite{he2016deep} backbone to extract multi-scale feature maps from each input view. To enhance multi-view consistency, these features are refined via a cross-view interaction module implemented with local window attention~\cite{liu2021swin}, which efficiently aggregates information from neighboring views.
Following feature extraction, the architecture proceeds to the multi-view depth prediction module. This module constructs a cost volume using a plane-sweep strategy with 128 inverse-depth candidates per reference view. By performing local neighbor matching on the cost volume, the network predicts robust depth maps that serve as the geometric basis for lifting 2D features into 3D space.
The lifted 3D representation is built in world coordinates and processed as a sparse voxel set. Specifically, we employ a sparse data structure where only occupied cells are materialized, avoiding the computational redundancy of a dense global grid. This sparse 3D refinement is efficiently implemented using MinkowskiEngine~\cite{choy20194d}.
In the default configuration, each occupied voxel predicts one Gaussian primitive with a 38-dimensional parameter vector, including opacity, center offset, anisotropic scale, quaternion rotation, and degree-2 spherical-harmonic color coefficients. This design ensures that primitive allocation remains scene-adaptive while preserving stable sparse 3D decoding.

\boldstart{More training details.}
Optimization uses AdamW~\cite{loshchilov2017decoupled} with decoupled weight decay, together with two learning-rate groups for the pretrained monocular branch and the remaining trainable parameters, and radient clipping is enabled for stability.
During training, each sample uses 6 input views and 8 target views.
Two input views are first selected as boundary anchors with a randomly sampled frame gap in a predefined range, and the remaining input views are sampled between these anchors.
The anchor-gap range is progressively expanded during early iterations.
All experiments are run at $256\times256$ input resolution.
Training is performed on RealEstate10K~\cite{zhou2018stereo} and then fine-tuned on ScanNet~\cite{dai2017scannet} from the RealEstate10K checkpoint, while ACID~\cite{liu2021infinite} is evaluated in a zero-shot manner.
For details on training objectives and weights, please refer to ~\cref{sec:training}.

\boldstart{Evaluation.}
Evaluation follows a controlled protocol for fair comparison with prior feed-forward 3DGS methods~\cite{charatan2024pixelsplat,chen2024mvsplat,kim2025transplat,xu2025depthsplat,zhang2024gaussian,jiang2025anysplat,liu2025worldmirror}.
For each scene, input views are selected using fixed frame-gap rules, and target novel views are chosen from disjoint camera positions that are not included in the inputs.
In our setup, each sample is evaluated on 8 target novel views.

\boldstart{Open-source.}
Our source codes are provided in the supplementary material. We will open-source the complete codebase for VolSplat.

\section{More Experimental Analysis}
\label{sec:app_experiment}

\begin{table}[t]
\centering
\caption{\textbf{Efficiency comparison.} All methods are evaluated with 6 input views on RealEstate10K~\cite{zhou2018stereo} via a single NVIDIA H20 GPU. Our VolSplat tops all image-quality metrics while retaining competitive inference efficiency even though utilizing 3D features. Note that absolute runtimes may differ from original studies due to hardware discrepancies, but relative rankings ensure fair efficiency comparison.}
\label{tab:app_efficiency}
\setlength{\tabcolsep}{3.5pt}
\resizebox{\textwidth}{!}{
\begin{tabular}{lcccccc}
 \toprule
 Method & PSNR$\uparrow$ & SSIM$\uparrow$ & LPIPS$\downarrow$ & Memory (GB) & Infer time (s) \\
 \midrule
 pixelSplat & 28.95 & 0.900 & 0.163 & 36.82 & 0.579 \\
 MVSplat & 29.13 & 0.924 & 0.091 & 4.70 & 0.369 \\
 TranSplat & 29.62 & 0.928 & 0.084 & 3.96 & 1.002 \\
 DepthSplat & 30.52 & 0.931 & 0.079 & 8.00 & 0.513 \\
 GGN & 26.68 & 0.879 & 0.133 & 4.70 & 0.377 \\
 AnySplat & 19.05 & 0.576 & 0.305 & 3.57 & 0.332 \\
 World-Mirror & 24.86 & 0.819 & 0.079 & 8.05 & 0.375 \\
 \textbf{Ours} & \textbf{31.30} & \textbf{0.941} & \textbf{0.075} & 4.65 & 0.575 \\
 \bottomrule
\end{tabular}
}
\end{table}

\boldstart{Efficiency Analysis.}
We evaluate the computational efficiency of our method against state-of-the-art baselines on RealEstate10K~\cite{zhou2018stereo} as illustrated in ~\cref{tab:app_efficiency}. All metrics utilize a single NVIDIA H20 GPU with 6 input views to ensure same evaluation environment.
VolSplat achieves a superior balance between reconstruction quality and resource consumption despite incorporating explicit 3D feature processing. Our approach outperforms all competing approaches while maintaining a competitive inference latency of 0.575s. This runtime remains comparable to leading pixel-aligned approaches~\cite{xu2025depthsplat}, which demonstrates that our voxel-aligned architecture introduces negligible overhead relative to the significant performance gains.
Furthermore, VolSplat exhibits competitive memory consumption by requiring only 4.65GB of VRAM.  This footprint is substantially lower than heavy-weight baselines~\cite{charatan2024pixelsplat,liu2025worldmirror}. Our method remains on par with lightweight alternatives such as MVSplat and ensures practicality for deployment on consumer-grade hardware. While absolute runtime values may vary due to hardware discrepancies compared to original publications, the relative rankings presented here reflect a controlled and fair comparison.

\section{Limitation and Societal Impacts}
\label{sec:limit_scoietal}

\boldstart{Limitation analysis.}
Our current framework assumes a static scene assumption during the multi-view feature aggregation and voxel construction steps. Consequently, VolSplat struggle to reconstruct dynamic objects or changing environments, as the geometric consistency enforced by our cost volume and sparse 3D U-Net~\cite{cciccek20163d} relies on multi-view consistency. Moving elements in the scene can lead to artifacts such as ghosting or blurring in the rendered novel views.

\boldstart{Potential and negative societal impacts.}
VolSplat significantly advances the capability of feed-forward 3D reconstruction by decoupling geometry from input pixel resolution. It enables the generation of high-fidelity 3D assets from sparse multi-view images with superior geometric consistency, positioning VolSplat as a valuable tool for immersive applications such as virtual reality, gaming, and digital twin creation.

While the ability to produce photorealistic 3D reconstructions from limited data is beneficial for content creation, it is important to acknowledge the potential for misuse. The high fidelity of the generated scenes could be exploited to create deepfakes or unauthorized digital replicas of private spaces. Consequently, the deployment of VolSplat in sensitive contexts should be accompanied by robust watermarking techniques or authentication protocols to mitigate the risks associated with the synthesis of misleading or non-consensual 3D content.

\section{More Visual Comparisons}
\label{sec:app_visual}

\begin{figure}[t]
\centering
\includegraphics[width=\linewidth]{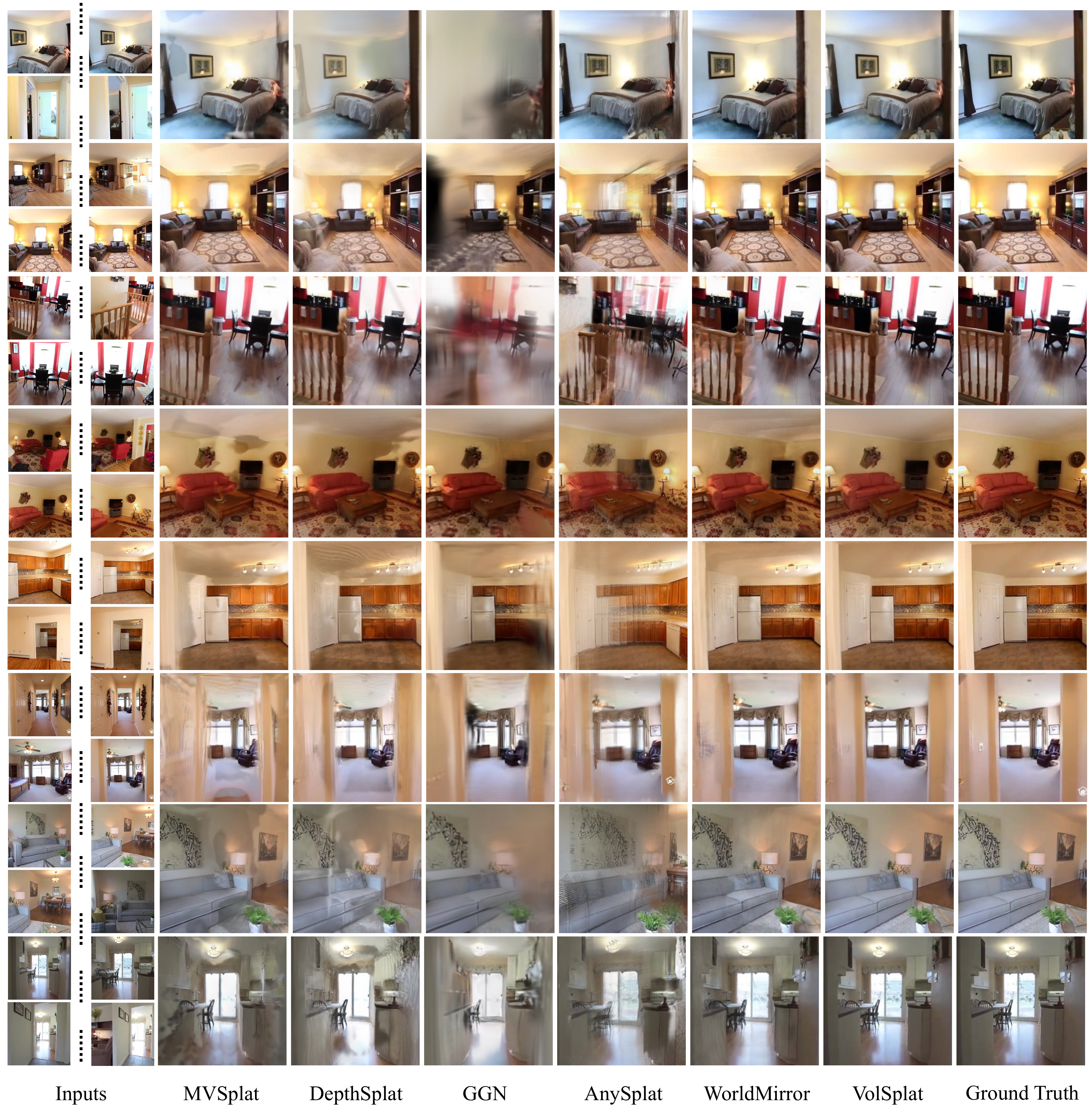}
\caption{\textbf{More qualitative comparisons on RealEstate10K~\cite{zhou2018stereo} under 6 input views.}}
\label{fig:re10k_6views}
\end{figure}

\begin{figure}[t]
\centering
\includegraphics[width=\linewidth]{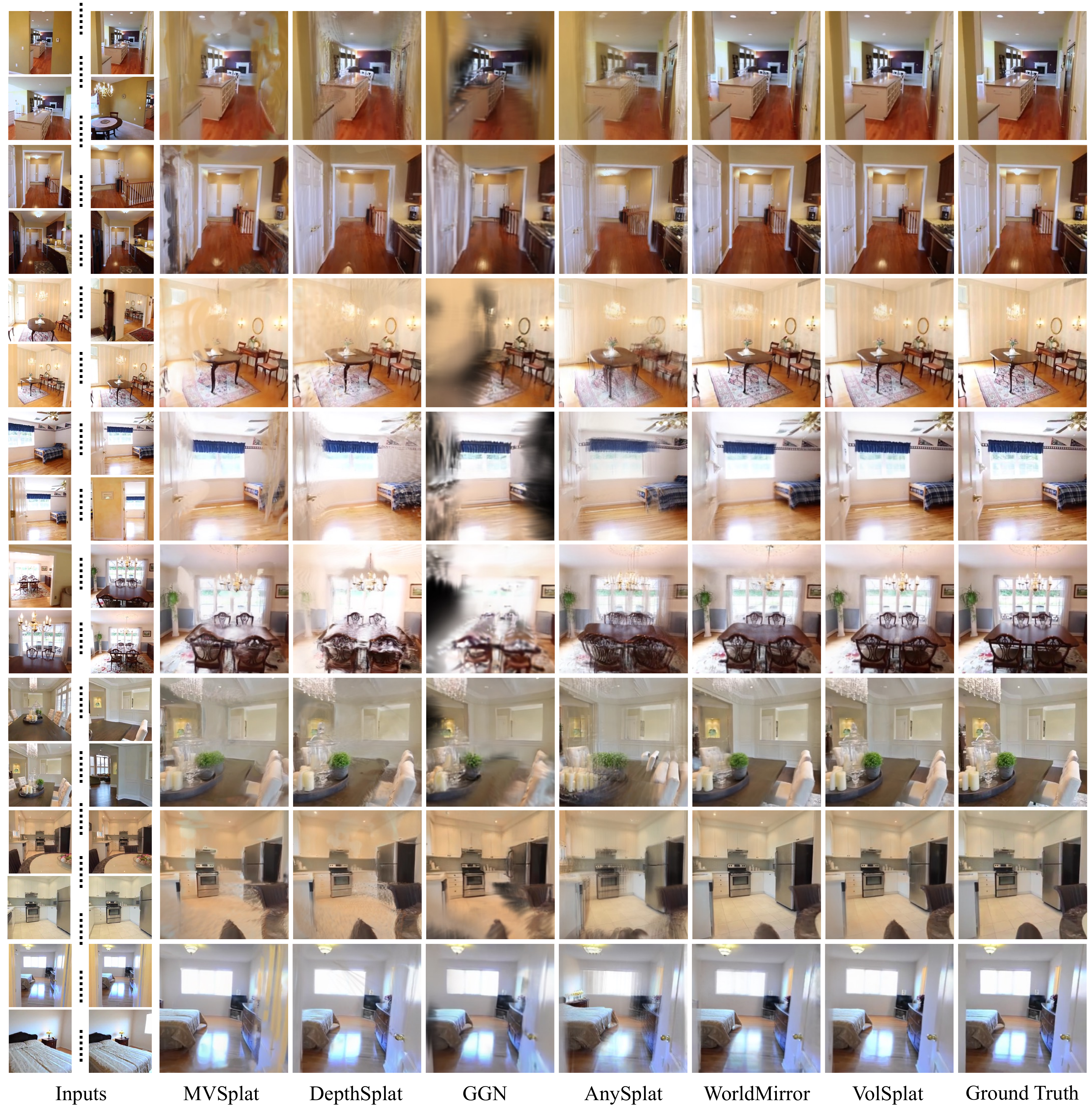}
\caption{\textbf{More qualitative comparisons on RealEstate10K~\cite{zhou2018stereo} under 12 input views.}}
\label{fig:re10k_12views}
\end{figure}

\begin{figure}[t]
\centering
\includegraphics[width=\linewidth]{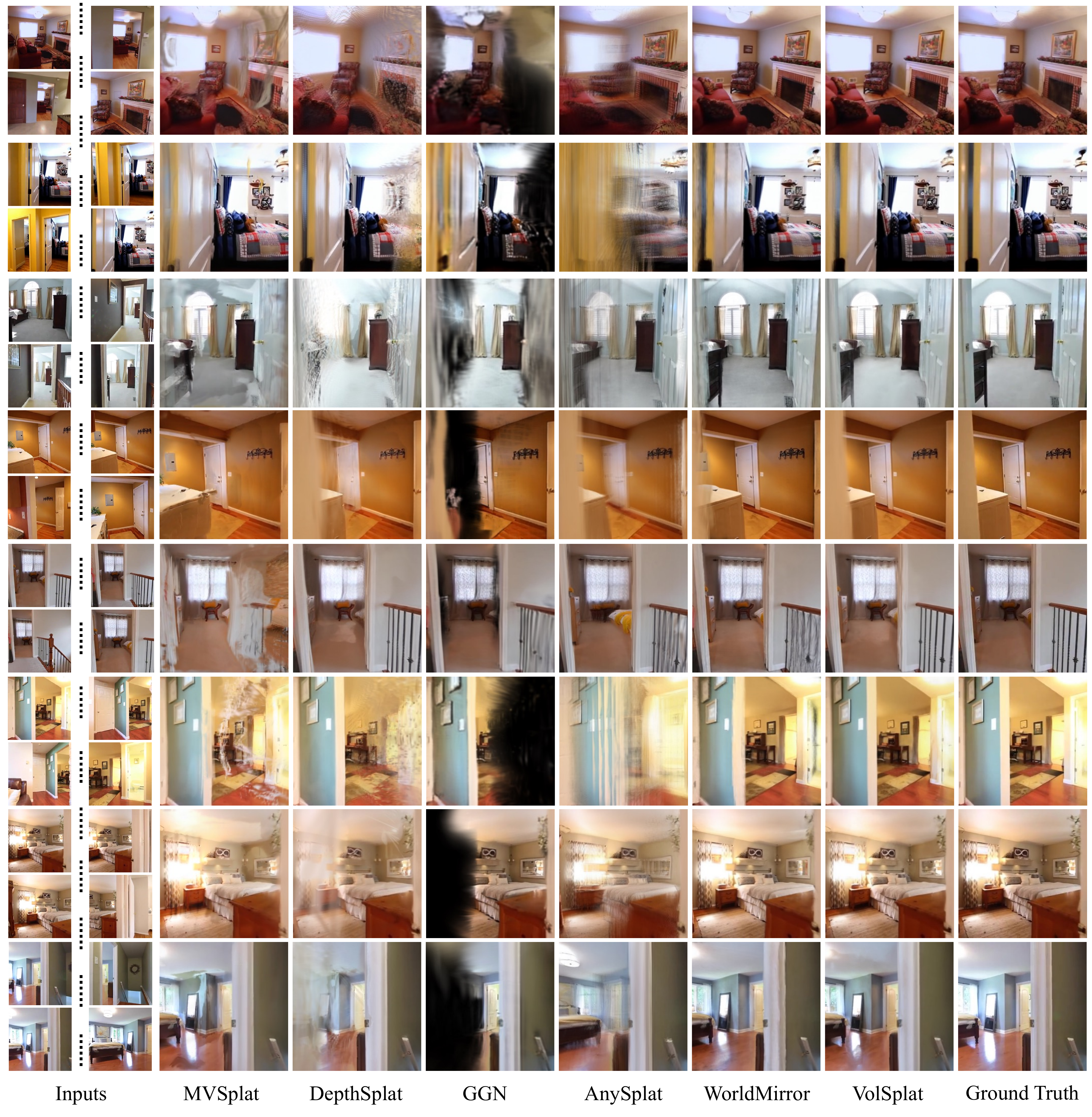}
\caption{\textbf{More qualitative comparisons on RealEstate10K~\cite{zhou2018stereo} under 24 input views.}}
\label{fig:re10k_24views}
\end{figure}

This section provides additional qualitative comparison results. We present further visualizations for VolSplat on the RealEstate10K~\cite{zhou2018stereo} dataset, comparing against SOTA baselines~\cite{charatan2024pixelsplat,chen2024mvsplat,kim2025transplat,xu2025depthsplat,zhang2024gaussian,jiang2025anysplat,liu2025worldmirror}.
To illustrate how VolSplat performs with varying numbers of input views, we showcase comparative results across different settings. For the standard setting, comparisons with 6 input views are presented in \cref{fig:re10k_6views}. To demonstrate scalability to denser inputs, visual comparisons between our method and competing baselines are displayed for scenarios with 12 and 24 input views in \cref{fig:re10k_12views} and \cref{fig:re10k_24views}, respectively. The corresponding quantitative results for these multi-view experiments can be found in ~\cref{tab:re10k_multiview}.

\end{document}